%% file: bmvc_final.tex
\documentclass{bmvc2k}

\title{DepthP+P: Metric Accurate Monocular Depth Estimation using Planar and Parallax}

\addauthor{Sadra Safadoust}{ssafadoust20@ku.edu.tr}{1}
\addauthor{Fatma Güney}{fguney@ku.edu.tr}{1}
\addinstitution{
 KUIS AI Center\\
 Koç University\\
 Istanbul, Turkey
}

\runninghead{Sadra Safadoust, Fatma Güney}{DepthP+P}

\def\eg{\emph{e.g}\bmvaOneDot}
\def\Eg{\emph{E.g}\bmvaOneDot}
\def\etal{\emph{et al}\bmvaOneDot}

\usepackage{graphicx}
\usepackage{adjustbox}
\usepackage{amsmath}
\usepackage{amssymb}
\usepackage{booktabs}
\usepackage{pifont}
\usepackage{multirow}

\begin{document}

\maketitle

\input{commands}

\input{sec/abstract}
\input{sec/intro}
\input{sec/rw}

\input{sec/methodology}
\input{sec/experiments}

\input{sec/conclusion}

\bibliography{bibliography_long, egbib}
\clearpage

\title{Supplementary Material for \\ DepthP+P: Metric Accurate Monocular Depth Estimation using Planar and Parallax}
\bmvaResetAuthors
\addauthor{Sadra Safadoust}{ssafadoust20@ku.edu.tr}{1}
\addauthor{Fatma Güney}{fguney@ku.edu.tr}{1}
\addinstitution{
 KUIS AI Center\\
 Koç University\\
 Istanbul, Turkey
}
\maketitlesecond

\def\thesection {\Alph{section}}

\setcounter{section}{0}
\renewcommand{\theHsection}{Supplement.\thesection}
\input{supp/sec/summary}
\input{supp/sec/derivation}

\input{supp/sec/normal}
\input{supp/sec/qual}

\end{document}

%% file: commands.tex
\newcommand{\Perp}{\perp\!\!\! \perp}
\newcommand{\bK}{\mathbf{K}}
\newcommand{\bX}{\mathbf{X}}
\newcommand{\bY}{\mathbf{Y}}
\newcommand{\bk}{\mathbf{k}}
\newcommand{\bx}{\mathbf{x}}
\newcommand{\by}{\mathbf{y}}
\newcommand{\bhy}{\hat{\mathbf{y}}}
\newcommand{\bty}{\tilde{\mathbf{y}}}
\newcommand{\bG}{\mathbf{G}}
\newcommand{\bI}{\mathbf{I}}
\newcommand{\bg}{\mathbf{g}}
\newcommand{\bS}{\mathbf{S}}
\newcommand{\bs}{\mathbf{s}}
\newcommand{\bN}{\mathbf{N}}
\newcommand{\bM}{\mathbf{M}}
\newcommand{\bw}{\mathbf{w}}
\newcommand{\eye}{\mathbf{I}}
\newcommand{\bU}{\mathbf{U}}
\newcommand{\bV}{\mathbf{V}}
\newcommand{\bW}{\mathbf{W}}
\newcommand{\bn}{\mathbf{n}}
\newcommand{\bv}{\mathbf{v}}
\newcommand{\bwv}{\mathbf{wv}}
\newcommand{\bq}{\mathbf{q}}
\newcommand{\bR}{\mathbf{R}}
\newcommand{\bi}{\mathbf{i}}
\newcommand{\bj}{\mathbf{j}}
\newcommand{\bp}{\mathbf{p}}
\newcommand{\bt}{\mathbf{t}}
\newcommand{\bJ}{\mathbf{J}}
\newcommand{\bu}{\mathbf{u}}
\newcommand{\bB}{\mathbf{B}}
\newcommand{\bD}{\mathbf{D}}
\newcommand{\bz}{\mathbf{z}}
\newcommand{\bP}{\mathbf{P}}
\newcommand{\bC}{\mathbf{C}}
\newcommand{\bA}{\mathbf{A}}
\newcommand{\bZ}{\mathbf{Z}}
\newcommand{\bff}{\mathbf{f}}
\newcommand{\bF}{\mathbf{F}}
\newcommand{\bo}{\mathbf{o}}
\newcommand{\bO}{\mathbf{O}}
\newcommand{\bc}{\mathbf{c}}
\newcommand{\bm}{\mathbf{m}}
\newcommand{\bT}{\mathbf{T}}
\newcommand{\bQ}{\mathbf{Q}}
\newcommand{\bL}{\mathbf{L}}
\newcommand{\bl}{\mathbf{l}}
\newcommand{\ba}{\mathbf{a}}
\newcommand{\bE}{\mathbf{E}}
\newcommand{\bH}{\mathbf{H}}
\newcommand{\bd}{\mathbf{d}}
\newcommand{\br}{\mathbf{r}}
\newcommand{\be}{\mathbf{e}}
\newcommand{\bb}{\mathbf{b}}
\newcommand{\bh}{\mathbf{h}}
\newcommand{\bhh}{\hat{\mathbf{h}}}
\newcommand{\btheta}{\boldsymbol{\theta}}
\newcommand{\bTheta}{\boldsymbol{\Theta}}
\newcommand{\bpi}{\boldsymbol{\pi}}
\newcommand{\bPi}{\boldsymbol{\Pi}}
\newcommand{\bphi}{\boldsymbol{\phi}}
\newcommand{\bPhi}{\boldsymbol{\Phi}}
\newcommand{\bmu}{\boldsymbol{\mu}}
\newcommand{\bSigma}{\boldsymbol{\Sigma}}
\newcommand{\bGamma}{\boldsymbol{\Gamma}}
\newcommand{\bbeta}{\boldsymbol{\beta}}
\newcommand{\bomega}{\boldsymbol{\omega}}
\newcommand{\blambda}{\boldsymbol{\lambda}}
\newcommand{\bLambda}{\boldsymbol{\Lambda}}
\newcommand{\bkappa}{\boldsymbol{\kappa}}
\newcommand{\btau}{\boldsymbol{\tau}}
\newcommand{\balpha}{\boldsymbol{\alpha}}
\newcommand{\nR}{\mathbb{R}}
\newcommand{\nN}{\mathbb{N}}
\newcommand{\nL}{\mathbb{L}}
\newcommand{\cN}{\mathcal{N}}
\newcommand{\cM}{\mathcal{M}}
\newcommand{\cR}{\mathcal{R}}
\newcommand{\cB}{\mathcal{B}}
\newcommand{\cL}{\mathcal{L}}
\newcommand{\cH}{\mathcal{H}}
\newcommand{\cS}{\mathcal{S}}
\newcommand{\cT}{\mathcal{T}}
\newcommand{\cO}{\mathcal{O}}
\newcommand{\cC}{\mathcal{C}}
\newcommand{\cP}{\mathcal{P}}
\newcommand{\cE}{\mathcal{E}}
\newcommand{\cI}{\mathcal{I}}
\newcommand{\cF}{\mathcal{F}}
\newcommand{\cK}{\mathcal{K}}
\newcommand{\cY}{\mathcal{Y}}
\newcommand{\cX}{\mathcal{X}}
\def\bgamma{\boldsymbol\gamma}

\newcommand{\specialcell}[2][c]{%
  \begin{tabular}[#1]{@{}c@{}}#2\end{tabular}}

\newcommand{\figref}[1]{\Fig~\ref{#1}}
\newcommand{\secref}[1]{Section~\ref{#1}}
\newcommand{\algref}[1]{Algorithm~\ref{#1}}
\newcommand{\eqnref}[1]{Eq.~\eqref{#1}}
\newcommand{\tabref}[1]{Table~\ref{#1}}

\newcommand{\rulesep}{\unskip\ \vrule\ }



\newcommand{\KLD}[2]{D_{\mathrm{KL}} \left( \left. \left. #1 \right|\right| #2 \right) }

\renewcommand{\b}{\ensuremath{\mathbf}}

\def\mc{\mathcal}
\def\mb{\mathbf}

\newcommand{\T}{^{\raisemath{-1pt}{\mathsf{T}}}}

\makeatletter
\DeclareRobustCommand\onedot{\futurelet\@let@token\@onedot}
\def\@onedot{\ifx\@let@token.\else.\null\fi\xspace}
\def\eg{e.g\onedot} \def\Eg{E.g\onedot}
\def\ie{i.e\onedot} \def\Ie{I.e\onedot}
\def\cf{cf\onedot} \def\Cf{Cf\onedot}
\def\etc{etc\onedot} \def\vs{vs\onedot}
\def\wrt{wrt\onedot}
\def\dof{d.o.f\onedot}
\def\etal{et~al\onedot} \def\iid{i.i.d\onedot}
\def\Fig{Fig\onedot} \def\Eqn{Eqn\onedot} \def\Sec{Sec\onedot} \def\Alg{Alg\onedot}
\makeatother

\newcommand{\xdownarrow}[1]{%
  {\left\downarrow\vbox to #1{}\right.\kern-\nulldelimiterspace}
}

\newcommand{\xuparrow}[1]{%
  {\left\uparrow\vbox to #1{}\right.\kern-\nulldelimiterspace}
}

\renewcommand\UrlFont{\color{blue}\rmfamily}

\newcommand*\rot{\rotatebox{90}}
\newcommand{\boldparagraph}[1]{\vspace{0.2cm}\noindent{\bf #1:} }

\newcommand{\sadra}[1]{ \noindent {\color{blue} {\bf Sadra:} {#1}} }
\newcommand{\ftm}[1]{ \noindent {\color{magenta} {\bf Fatma:} {#1}} }

%% file: sec/abstract.tex
\begin{abstract}
Current self-supervised monocular depth estimation methods are mostly based on estimating a rigid-body motion representing camera motion.
These methods suffer from the well-known scale ambiguity problem in their predictions. 
We propose DepthP+P, a method that learns to estimate outputs in metric scale by following the traditional planar parallax paradigm. 
We first align the two frames using a common ground plane which removes the effect of the rotation component in the camera motion.
With two neural networks, we predict the depth and the camera translation, which is easier to predict alone compared to predicting it together with rotation. 
By assuming a known camera height, we can then calculate the induced 2D image motion of a 3D point and use it for reconstructing the target image in a self-supervised monocular approach. 
We perform experiments on the KITTI driving dataset and show that the planar parallax approach, which only needs to predict camera translation, can be a metrically accurate alternative to the current methods that rely on estimating 6DoF camera motion. 
\end{abstract}

%% file: sec/intro.tex
\section{Introduction}
\label{sec:introduction}
Understanding the 3D structure of a scene is fairly easy for human beings. We can easily reason about our surroundings and decompose them into different objects. Having this ability is crucial for autonomous vehicles to be able to drive in different environments. Training deep networks for estimating depth has proven successful in computer vision research. However, many such methods are supervised and require ground truth depth which is costly to achieve. Another line of work uses a stereo setup that must be carefully calibrated. Both of these approaches cannot use the vast amount of unlabeled videos that are easily available for training. On the other hand, self-supervised monocular depth estimation methods that do not rely on stereo supervision do not suffer from these limitations and, in practice, have been closing the gap with their supervised or stereo counterparts. 

Current self-supervised monocular depth estimation approaches all follow the same basic idea~ proposed in~\cite{Zhou2017CVPR}. They use a pose network to estimate the ego-motion between a source frame and the target frame and a depth network to estimate the depth of the target image. These estimations can then be used to sample pixels from the source image to synthesize the target frame. The difference between the target frame and the synthesized can be used as the source of supervision for training the networks. In this paper, we propose another approach to synthesize the target image. Our approach, \textbf{DepthP+P}, illustrated in \figref{fig:pipeline}, uses the traditional planar parallax formulation~\cite{Sawhney1994CVPR, Irani1996ECCV}, which decomposes the motion into a planar homography and a residual parallax. Consider a plane in the scene and its motion represented by a homography from the source to the target image. By first warping the source image according to this homography, the motion of the plane is canceled. Then the residual image motion depends on two factors: (1) the deviations of the scene structure from the plane, \ie the depth of points and their perpendicular distance to the plane and (2) only the translational motion of the camera. Autonomous driving is a perfect use case for this approach because there is typically a planar surface in front of the vehicle, \ie the road. However, it is important to note that the plane in the planar parallax formulation does not necessarily have to be a real plane and can also be a virtual plane, but choosing the road as the planar surface makes it easier to implement in practice. Moreover, our approach does not rely on the availability of a plane to predict depth during inference.

\input{figures/pipeline}
In this approach, we first align the road plane between the source and target images. This is achieved by calculating the homography between the road regions in two frames and then warping the source frame according to the homography to obtain the aligned image. By doing so, the road regions in the aligned image and target image match. 
The residual motion between the aligned image and the target image can be explained as follows: We first estimate the depth of each pixel with a monocular depth network and back-project them into 3D. Then using a known camera height, we can calculate the perpendicular distance of each point to the road. In addition, we estimate the translation between the camera origins. Note that this is different from the typical monocular depth approach \cite{Zhou2017CVPR, Godard2019ICCV} which needs to estimate both the rotation and translation components. Finally, the target image can be synthesized from the aligned image using the calculated residual parallax as shown in \figref{fig:pp}.

The planar parallax approach for self-supervised monocular estimations has a number of advantages over the previous paradigm. Firstly, it is much easier to optimize because it removes the ambiguities associated with predicting rotational camera motion~\cite{Irani2002TPAMI}. Secondly, it can produce metric accurate outputs. Previous monocular depth methods can estimate depth and motion up to a scale. Typically, during inference, ground truth depth data is used to scale the predicted depth values such that the median of the predicted depth is equal to that of ground truth depth~\cite{Zhou2017CVPR}. Our approach is able to predict metric accurate depth without needing ground truth depth data by only assuming a known camera height. 

%% file: figures/pipeline.tex
\begin{figure*}[t]
    \includegraphics[width=\linewidth]{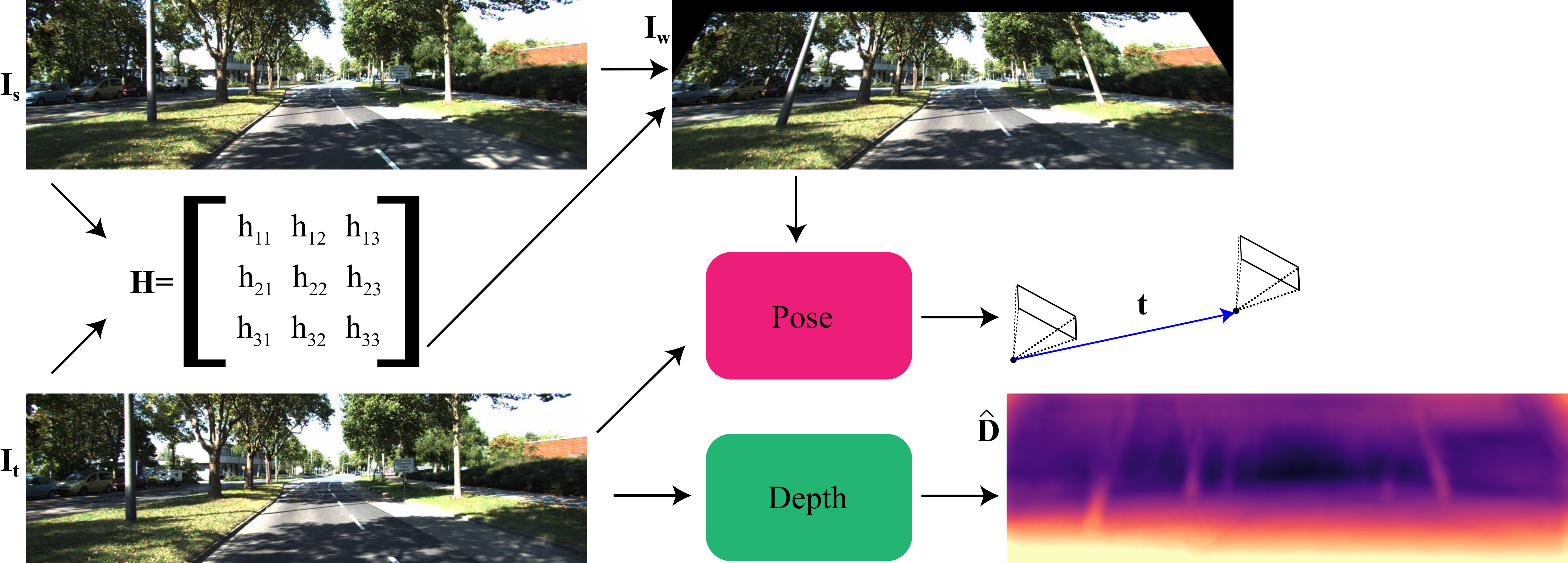}
    \caption{\textbf{Overview of our Approach.} Using the source image $\bI_s$ and the target image $\bI_t$, we first calculate the homography $\bH$ that aligns the road plane across these two images. We then warp $\bI_s$ according to $\bH$ and obtain the aligned image $\bI_w$. The aligned image $\bI_w$ and the target image $\bI_t$ are input to the pose network which estimates the camera translation $\bt$ only. The depth network takes the $\bI_t$ and produces a metric accurate depth map $\hat{\bD}$.
    }
    \label{fig:pipeline}
    \vspace{-.5cm}
\end{figure*}

%% file: sec/rw.tex
\section{Related Work}
\label{sec:related_work}
\subsection{Self-Supervised Monocular Depth}
\boldparagraph{View Synthesis} Garg \etal~\cite{Garg2016ECCV} were the first to propose a method that uses view synthesis as an objective for depth estimation from single images. 
Monodepth~\cite{Godard2017CVPR} uses Spatial Transformer Networks~(STNs)~\cite{Jaderberg2015NeurIPS} to synthesize the images in a fully-differentiable way. 
SfmLearner~\cite{Zhou2017CVPR} generalizes view synthesis to temporally consecutive images by using another network to predict the relative pose between them. 
Zhan \etal~\cite{Zhan2018CVPR} use stereo sequences to perform view synthesis using temporally consecutive pairs as well as the left-right pairs, enabling them to benefit from both monocular and stereo supervision. In addition to image reconstruction, they also use feature reconstruction as supervision. Similarly, by going beyond pixel-wise reconstruction error, Mahjourian \etal~\cite{Mahjourian2018CVPR} propose to use a 3D point cloud alignment loss to enforce the estimated point clouds and the camera pose to be consistent temporally. Wang \etal~\cite{Wang2018CVPR} use direct visual odometry in a differentiable manner to solve for ego-motion using the estimated depth. 

In addition to depth and camera pose, several methods estimate optical flow for residual motion. After predicting the camera motion, GeoNet~\cite{Yin2018CVPR} estimates the remaining object motion using optical flow. In order to prevent the errors of camera pose or depth predictions from propagating to flow estimations, DF-Net~\cite{Zou2018ECCV} enforces consistency between optical flow and the flow induced by the depth and pose predictions. GLNet~\cite{Chen2019ICCV} uses epipolar constraint for optical flow, along with other geometric constraints, further improving the performance. 
EPC++~\cite{Luo2019PAMI} proposes a holistic 3D motion parser that uses predicted depth, pose, and optical flow to estimate segmentation masks for dynamic objects and their motion as well as background motion. Ranjan \etal~\cite{Ranjan2019CVPR} jointly train networks for depth, pose, optical flow, and motion segmentation so that they can use geometric constraints on the static regions and generic optical flow on moving objects. MonoDepthSeg~\cite{Safa2021ThreeDV} proposes to jointly estimate depth, independently moving regions, and their motion with an efficient architecture.

Some approaches keep the original framework with a depth and a pose network but improve the performance with better loss functions, improved network architectures, and innovative design choices. When estimating depth at multiple scales, Monodepth2~\cite{Godard2019ICCV} proposes to first upsample the estimated low-scale depths to the input image size and then calculate the photometric loss at that scale. 
Monodepth2 also proposes to calculate the minimum of reprojection errors per pixel instead of averaging them when synthesizing the target image from multiple views to prevent blurry depth estimations. PackNet~\cite{Guizilini2020CVPR} changes the architecture of the depth network and uses 3D convolutions to learn to preserve spatial information using symmetrical 3D packing and unpacking blocks for predicting depth.

\boldparagraph{Scale Ambiguity} Self-supervised monocular depth estimation models suffer from the scale ambiguity problem, and the depth and pose outputs of such models are in an unknown scale. The median scaling technique used by many previous methods does not actually solve this problem because it relies on ground truth depth data during inference which is not always easily available. Bian \etal~\cite{Bian2019NeurIPS} introduce a loss to minimize normalized differences of depth maps across the entire sequence. This makes the estimations globally scale-consistent. However, although this means that the predictions are at the same scale, that specific scale is still unknown, and the median scaling is still required during evaluation. 

There are a number of monocular methods that can output depth estimations in absolute scale. Roussel \etal~\cite{Roussel2019IROS} use a network that was pre-trained with stereo pairs on a dataset and finetunes it on another dataset while maintaining the metric scale. Guizilini \etal~\cite{Guizilini2020CVPR} propose a version of their PackNet that uses ground truth camera velocity and the timestamps of images to enforce the estimations to be metrically accurate. Bartoccioni \etal~\cite{Bartoccioni2021Arxiv} supervise their depth predictions with a sparse LiDAR. However, all of these approaches rely on ground truth data from extra sensors during training. 

There are a few other methods that do not require additional supervision and only use the camera height to achieve depth estimations in metric units similar to the proposed method. DNet~\cite{Xue2020IROS} estimates the ground plane during inference and, using the real height of the camera, recovers the scale of the predictions. However, it needs a ground plane to be visible during the test time. In other words, they do not train their depth outputs to be in absolute scale. Rather, they recover the scale of the estimations with another module during test time. Wagstaff and Kelly~\cite{Wagstaff2021IROS} train a network that learns the metric scale during training using camera height. They introduce a plane segmentation network and propose a three-staged training procedure for training the depth estimation model in metric scale. First, they train an unscaled depth network and then use it to train the plane segmentation network. Finally, they train a new metrically accurate depth network using the pre-trained plane segmentation network.
Similar to~\cite{Wagstaff2021IROS}, we also learn the metric scale during training, but we do not need a multi-stage process, nor do we rely on the existence of a ground plane during inference, differently from previous work~\cite{Xue2020IROS}.

\subsection{Planar Parallax}
The Planar Parallax paradigm, also called Plane + Parallax (P+P), has been used to understand the 3D structure of a scene from multiple images by decomposing the motion into a planar homography and a residual parallax. Sawhney~\cite{Sawhney1994CVPR} proposes a formulation for the residual parallax that uses depth and distance to the plane. Irani \etal~\cite{Irani1996ECCV} use this formulation to derive a rigidity constraint between pairs of points over multiple images. 
Irani \etal~\cite{Irani1998ECCV} derive trifocal constraints and use them to propose a simple method for new view synthesis. In a follow-up work~\cite{Irani2002TPAMI}, they extend the planar parallax method to more than two uncalibrated frames. 

More recently, MR-Flow~\cite{Wulff2017CVPR} uses P+P to refine the optical flow estimations with rigidity constraints. Chaney \etal~\cite{Chaney2019IROS} use P+P to estimate the height of points in the scene with event-based cameras.
We propose a method to use the P+P formulation within the view synthesis framework for self-supervised monocular depth estimation.

%% file: sec/methodology.tex
\section{Methodology}
\label{sec:methodology}
\input{figures/pp}
Despite the success of current self-supervised monocular depth estimation approaches, they suffer from scale ambiguity. \ie the estimated depth values are in an unknown scale. Therefore, in order to evaluate and compare these methods, they are usually normalized using the median scaling approach~\cite{Zhou2017CVPR}. Here, we propose an approach that predicts depth maps in metric scale without using any ground truth depth supervision. 

\subsection{DepthP+P} 
Our approach is based on the Planar Parallax decomposition which has been studied in detail before \cite{Sawhney1994CVPR, Irani1996ECCV}. We first introduce it here to establish our notation and then build our method to predict depth following that notation.

\boldparagraph{Notation} Let $\bPi$ be a 3D plane and $\bH$ be the homography aligning $\bPi$ between the target image $\bI_t$ and the source image $\bI_s$.
Let $\bp$ and $\bp'$ be the images of the 3D point $\bx=[\bX,\bY,\bZ]^T$ on the $\bI_t$ and $\bI_s$ respectively. As shown on the left in \figref{fig:pp}, we can warp $\bp'$ by the homography $\bH$ and obtain the image point $\bp_w$:
\begin{equation}
    \bp_w \sim \bH \bp'
\end{equation}
where we omit the conversion to the homogenous coordinates. Note that by warping the source image $\bI_s$, we obtain the aligned image $\bI_w$ such that the plane $\bPi$ matches between them. The displacement between $\bp_w$ and $\bp$ can be computed as follows:
\begin{equation}
    \label{eq:parallax}
    \bp_w - \bp = \frac{\bgamma}{\bd_c - \bgamma \bt_z}(\bt_z \bp - \bK \bt)
\end{equation}
where $\bK$ is the camera intrinsic, $\bt=[\bt_x, \bt_y, \bt_z]^T$ is the translation vector between the $\bI_t$ and $\bI_s$, and $\bd_c$ is the distance between the camera for the source view to the plane $\bPi$. The structure is represented by $\bgamma = \frac{\bh}{\bZ}$ where $\bh$ is the distance of $\bx$ to $\bPi$.
Note that when $\bx$ lies on the plane $\bPi$, \ie $\bh=0$, we will have $\bp_w = \bp$.

\boldparagraph{DepthP+P} Following the typical self-supervised monocular depth approach, our framework has two networks, one for estimating depth and another for estimating the translation between frames. Note that, unlike other methods, we do not need to estimate the rotation between the two views.
Precisely, our pose network takes the source and images $\bI_s, \bI_w$  and outputs the translation vector $\bt$. The depth network takes the target image $\bI_t$ and outputs the depth map $\hat{\bD}$ for  $\bI_t$. 

For every pixel $\bp=[x,y]$, let $\hat{\bD}{(\bp)}$ denote its estimated depth. We backproject $\bp$ using the camera intrinsics and the estimated depth to obtain the corresponding 3D point $\hat{\bx}$ in the camera coordinate system as follows:
\begin{equation}
    \label{eq:backproject}
    \hat{\bx} = \hat{\bD}{(\bp)}~\bK^{-1}~[x, y, 1]^T.
\end{equation}
%
Therefore, as demonstrated on the right in \figref{fig:pp}, we  have the following:
\begin{equation}
    \label{eq:height}
    \hat{\bh} = \bd_c - \bN^T \hat{\bx}, \quad  \quad \hat{\bgamma} = \frac{\hat{\bh}}{\hat{\bD}{(\bp)}}
\end{equation}
where $\bN$ is the normal vector of the plane $\bPi$, $\hat{\bh}$ is the estimated distance of the point $\hat{\bx}$ to the plane $\bPi$ and $\hat{\bgamma}$ is our estimate of the structure variable $\bgamma$. As a result, we obtain all the parameters required to use \eqnref{eq:parallax} to reconstruct the target image $\bI_t$ by warping the aligned image $\bI_w$ resulting in $\hat{\bI}_w$. In other words, for each pixel $\bp$ on the $\bI_t$, we  calculate the $\bp_w$ using \eqref{eq:parallax} according to the depth and translation predicted by our two networks and then inverse warp $\bI_w$ and obtain $\hat{\bI}_w$ to reconstruct $\bI_t$:
\begin{equation}
    \bI_t(\bp) \approx \hat{\bI}_w(\bp) = 
    \bI_w(\bp_w)
\end{equation}
We minimize the difference between $\bI_t$ and $\hat{\bI}_w$ for supervision as explained in the \secref{subsection:loss}.

In order to obtain the aligned images, we perform a pre-processing step on the dataset. We calculate a homography for every consecutive frame by using the road as the plane $\bPi$ and warp the frames according to the calculated homographies. In other words, we calculate a homography $\bH$ for every target image $\bI_t$ and source image $\bI_s$ pair, and then warp $\bI_s$ according to $\bH$ to obtain the warped source image $\bI_w$. We explain the details of this pre-processing step in \secref{subsection:dataset}.

\subsection{Self-Supervised Training Loss} \label{subsection:loss}
In our approach, we define our photometric loss function as the linear combination of the L1 distance and the structural similarity (SSIM) \cite{Wang2004TIP} to minimize the difference between the target image $\bI_t$ and the reconstructed image $\hat{\bI}_w$. Our photometric loss is therefore defined as follows: 
\begin{equation}
    \cL_{\textrm{photo}}(\bp) =   \left( 1-\alpha\right)\lvert \bI_t(\bp)-\hat{\bI}_w(\bp)\rvert + \frac{\alpha}{2}\left(1-\textrm{SSIM} (\bI_t, \hat{\bI}_w )(\bp) \right) 
\end{equation}
where we set $\alpha=0.85$. Note that for every target image we consider two aligned images. One from warping the previous frame, and one from warping the next frame. We use the per-pixel minimum reprojection error introduced in \cite{Godard2019ICCV} and calculate the minimum of the $\cL_{photo}$ for each pixel across the previous and next aligned images. 
We also define $\cL_{smooth}$ as an edge-aware smoothness loss over the mean-normalized inverse depth estimates \cite{Wang2018CVPR} to encourage the depth predictions to be locally smooth. Our total loss function is a combination of $\cL_{smooth}$ and $\cL_{photo}$ averaged over all $N$ pixels:
\begin{equation}
   \cL = \frac{1}{N} \sum_{\bp}  \lambda~ \cL_{smooth}(\bp) + \min_w (\cL_{photo}(\bp)) 
\end{equation}
where $\displaystyle \min_w$ calculates the minimum over the previous and next aligned frames and  $\lambda$ is a hyperparameter controlling the effect of the loss terms.

\subsection{Network Architecture}
Our depth network is based on the U-Net architecture \cite{Ronneberger2015MICCAI}. We use a ResNet \cite{He2016CVPR} pre-trained on the ImageNet \cite{Russakovsky2015IJCV} as the encoder for our depth network, and for the decoder we use the architecture similar to the one used by \cite{Godard2019ICCV}. The difference is that we directly estimate depth by multiplying the output of the last sigmoid layer by 250, which is the maximum depth value that can be predicted, instead of estimating disparity as in \cite{Godard2019ICCV}.  The depth network takes as input a single target image $\bI_t$ and outputs the per-pixel depth estimates. Note that the output of our depth decoder is in metric scale.

In DepthP+P, our second network takes $\bI_w$ and $\bI_t$ and outputs only the translation vector between the views. The network is similar to the pose network proposed in \cite{Godard2019ICCV}, except that the output is a 3-element vector representing the translation and is metric scale in our case. 

%% file: figures/pp.tex
\begin{figure*}[t]
    \includegraphics[width=0.49\linewidth]{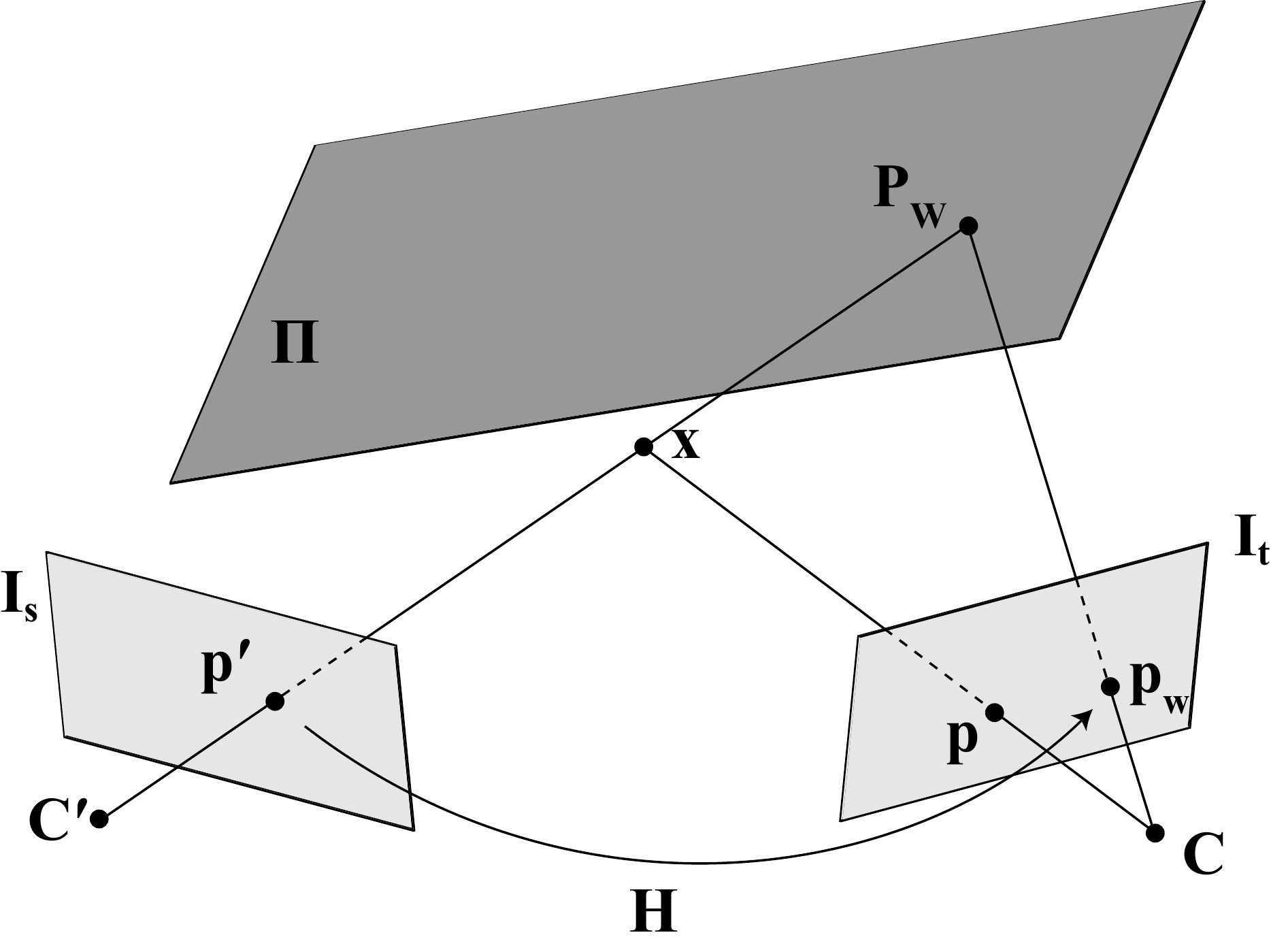}
    \hspace{0.02\linewidth}
    \raisebox{0.7cm}{\includegraphics[width=0.49\linewidth]{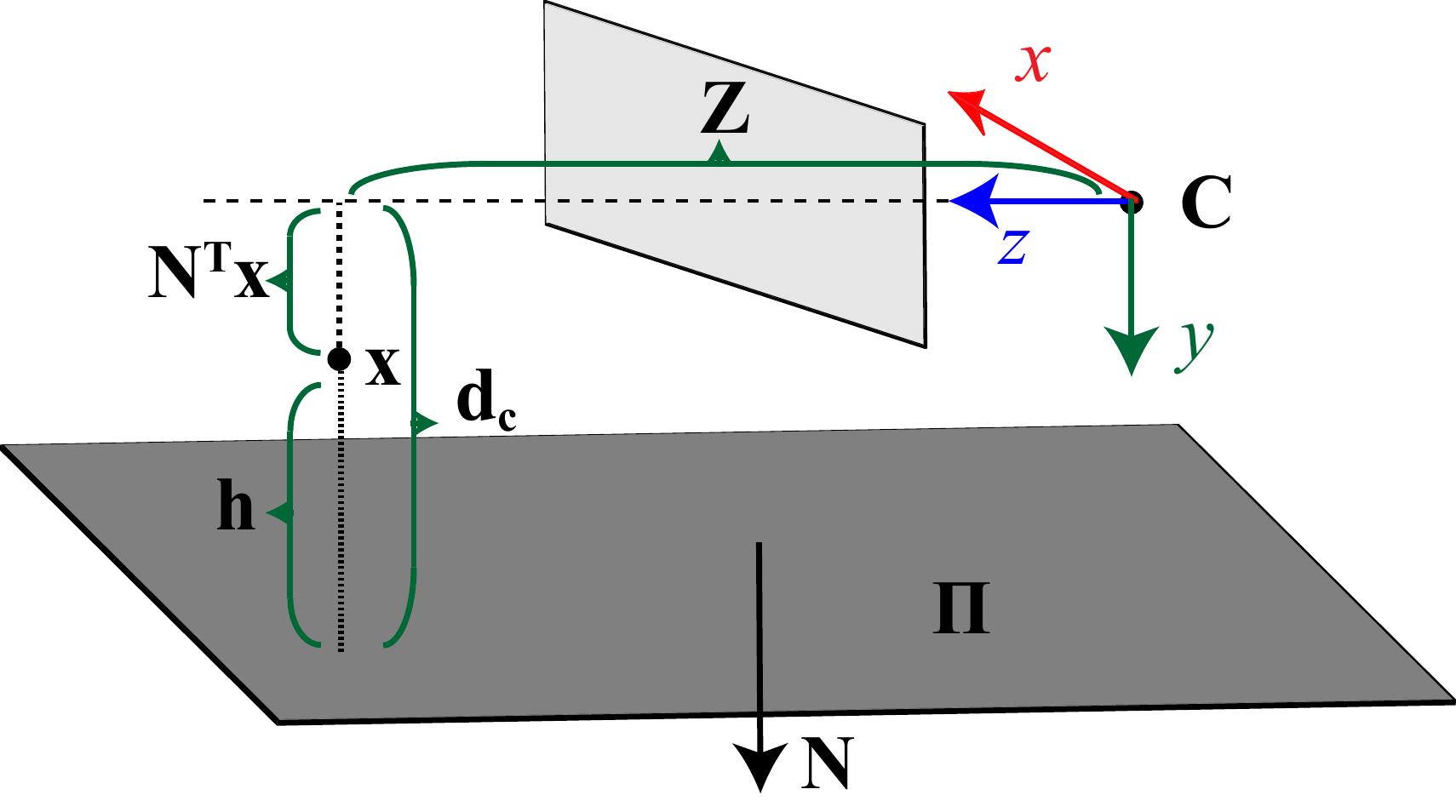}}
    \caption{\textbf{Visualization of the Planar Parallax.} \textit{Left:} The 3D point $\bx$ is projected to points $\bp$ and $\bp'$ on the target image $\bI_t$ and the source image $\bI_s$ respectively. Using the homography $\bH$ induced by the plane $\bPi$, the point $\bp'$ will be transformed to point $\bp_w$ on the target image. \textit{Right:} Calculating $\bh$, distance of $\bx$ to the $\bPi$ using the camera height $\bd_c$ and the normal vector $\bN$ of the plane. $\bC$ and $\bC'$ are the camera centers of $\bI_t$ and $\bI_s$ and $\bZ$ is the depth of the point.
    }
    \label{fig:pp}
    \vspace{-3.5pt}
\end{figure*}

%% file: sec/experiments.tex
\section{Experiments}
\label{sec:experiments}

\subsection{Dataset} \label{subsection:dataset}
\boldparagraph{KITTI} We use the Eigen split~\cite{Eigen2014NeurIPS} of the KITTI dataset~\cite{Geiger2013IJRR, Geiger2012CVPR} to train and evaluate our model. We use all of the images in the split for which we could accurately estimate the homography for aligning the road between consecutive images as explained in the next paragraph. This results in 45000 training and 1769 validation samples. We evaluate our model on the 697 test images in the split using the original ground truth provided by LiDAR. We also report results using the improved ground truth for 652 test images provided by Uhrig et al~\cite{Uhrig2017THREEDV}. They use a stereo-reconstruction method to remove the outliers in LiDAR points and increase the ground truth density by accumulating laser scans which result in high-quality ground truth data. The camera height in this dataset is $\bd_c=1.65$ and we assume that the road is completely horizontal, \ie $\bN = [0, 1, 0]^T$. 

\boldparagraph{Pre-processing the dataset for DepthP+P}
In order to use our P+P approach, we need to calculate the homography between the consecutive frames and warp the source frame according to the estimated homographies.   
Since we work on the driving scenarios on KITTI, we choose the ``road'' as our plane $\bPi$ which is visible in most of the frames. For calculating the homography, we need to find a set of (at least 4) corresponding pairs of road pixels between a source view $\bI_s$ and the target view $\bI_t$, \ie two consecutive images. For this purpose, we use the optical flow between $\bI_s$ and $\bI_t$ using \cite{Teed2020ECCV} to find the corresponding pixels. We then use \cite{Zhu2019CVPR} to select only the pixels that belong to the semantic class ``road''. Using the corresponding pairs of road pixels, we estimate the homography $\bH$ using OpenCV's RANSAC-based robust method. We do this to find the homography $\bH$ for all of the consecutive pairs of frames on KITTI. Note that for any consecutive pair of frames $\bI_1, \bI_2$ and the homography $\bH$ between them, we use $\bH$ to warp $\bI_1$ towards $\bI_2$ and also use $\bH^{-1}$ to warp $\bI_2$ towards $\bI_1$.  

\subsection{Depth Estimation Results}
\input{tables/kitti_main}
In \tabref{tab:kitti_main}, we report the depth estimation results of our method on the KITTI Eigen split using both the original and the improved ground truth. To the best of our knowledge, this is the first time that a deep learning model has been trained with view synthesis through the planar parallax paradigm (\eqnref{eq:parallax}). All of the previous methods are trained based on estimating the pose whereas our method introduces a novel approach. We can see that our method achieves significantly better results than the initial models by predicting the pose and depth. After the initial proposal of SfMLearner by Zhou \etal~\cite{Zhou2017CVPR}, several improvements have been proposed to improve its performance. Therefore, we believe that similar improvements can follow our model as future work to make it perform better than our initial proposal as well as the other state-of-the-art models that are trained to estimate the full pose. 

Note that \cite{Xue2020IROS} is not trained to estimate metrically accurate depth. Instead, its depth network outputs depth in an unknown scale, and then during inference, it needs a ground plane to be visible on the image to recover the scale of the network. When the ground plane is not visible on the image, \cite{Xue2020IROS} fails completely as shown in \figref{fig:noground}. As can be seen in this figure, this image from the KITTI dataset does not have a ground plane, and \cite{Xue2020IROS} cannot recover the scale and produces completely wrong estimates. While our method needs a ground plane during training, it does not rely on the availability of the ground plane during  inference, therefore it can still perform well. For reference, the absolute relative~(Abs Rel) error of \cite{Xue2020IROS} on \figref{fig:noground} is 1.178, while our model achieves a 0.252 error. \cite{Wagstaff2021IROS} achieves better results by using a pre-trained plane segmentation network in addition to the depth network, while our approach can achieve comparable results without a separate segmentation network.

\input{tables/stereo}

\input{figures/noground}

DepthP+P can also be trained with additional stereo supervision. In the proposed approach, we obtain monocular supervision from the P+P paradigm. In addition, using the known camera baseline and the estimated depth, we can warp the other image in the stereo setup to the input image for additional supervision signal. In \tabref{tab:stereo}, we report the performances of the methods that also use stereo supervision for training. Using stereo supervision significantly improves the performance of our DepthP+P model, outperforming all methods except for Monodepth2~\cite{Godard2019ICCV}. We show that by using a ResNet50 backbone instead of ResNet18, DepthP+P can obtain comparable results to Monodepth2~\cite{Godard2019ICCV}.

%% file: tables/kitti_main.tex
\begin{table*}[t]
    \centering
    \begin{adjustbox}{max width=\textwidth}
    \begin{tabular}{l | l | c | c c c c | c c c }
    \toprule
& {\multirow{2}{*}{\bf Method}} & {\multirow{2}{*}{\bf Scale}} & \multicolumn{4}{c}{Lower Better} & \multicolumn{3}{c}{Higher Better} \tabularnewline 
& & & Abs Rel & Sq Rel & RMSE & RMSE$_{log}$ & $\delta < 1.25$ & $\delta < 1.25^2$ & $\delta < 1.25^3$\tabularnewline \hline
\parbox[t]{2mm}{\multirow{16}{*}{\rotatebox[origin=c]{90}{\small{Original Ground Truth }}}}
& Zhou \etal \cite{Zhou2017CVPR} & \ding{55} & 0.183 & 1.595 & 6.709 & 0.270 & 0.734 & 0.902 & 0.959 \tabularnewline 
& Yang \etal \cite{Yang2018AAAI} & \ding{55} & 0.182 & 1.481 & 6.501 & 0.267 & 0.725 & 0.906 & 0.963 \tabularnewline 
& Mahjourian \etal \cite{Mahjourian2018CVPR} & \ding{55} & 0.163 & 1.240 & 6.220 & 0.250 & 0.762 & 0.916 & 0.968 \tabularnewline
& Yin \etal \cite{Yin2018CVPR} & \ding{55} & 0.149 & 1.060 & 5.567 & 2.226 & 0.796 & 0.935 & 0.975 \tabularnewline
& Wang \etal \cite{Wang2018CVPR} & \ding{55} & 0.151 & 1.257 & 5.583 & 0.228 & 0.810 & 0.936 & 0.974 \tabularnewline
& Zou \etal \cite{Zou2018ECCV} & \ding{55} & 0.150 & 1.124 & 5.507 & 0.223 & 0.806 & 0.933 & 0.973 \tabularnewline
& Yang \etal \cite{Yang2018CVPR} & \ding{55} & 0.162 & 1.352 & 6.276 & 0.252 & - & - & - \tabularnewline
& Ranjan \etal \cite{Ranjan2019CVPR} & \ding{55} & 0.148 & 1.149 & 5.464 & 0.226 & 0.815 & 0.935 & 0.973 \tabularnewline
& Luo \etal \cite{Luo2019PAMI} & \ding{55} & 0.141 & 1.029 & 5.350 & 0.216 & 0.816 & 0.941 & 0.976 \tabularnewline
& Chen \etal \cite{Chen2019ICCV} & \ding{55} & 0.135 & 1.070 & 5.230 & 0.210 & 0.841 & 0.948 & \underline{0.980} \tabularnewline
& Godard \etal \cite{Godard2019ICCV} & \ding{55} & \textbf{0.110} & 0.831 & \underline{4.642} & \textbf{0.187} & \textbf{0.883} & \textbf{0.962} & \textbf{0.982} \tabularnewline
& Guizilini \etal \cite{Guizilini2020CVPR} & \ding{55} & \underline{0.111} & \textbf{0.785}   & \textbf{4.601} & \underline{0.189} & 0.878 & \underline{0.960} & \textbf{0.982}\tabularnewline 
& Safadoust \etal \cite{Safa2021ThreeDV} & \ding{55} & \textbf{0.110} & \underline{0.792} & 4.700 & \underline{0.189} & \underline{0.881} & \underline{0.960} & \textbf{0.982} \tabularnewline
& Xue \etal \cite{Xue2020IROS} & \ding{51} & 0.118 & 0.925 & 4.918 & 0.199 & 0.862 & 0.953 & 0.979 \tabularnewline
& Wagstaff and Kelly \cite{Wagstaff2021IROS}& \ding{51} & 0.123 & 0.996 & 5.253 & 0.213 & 0.840 & 0.947 & 0.978 \tabularnewline
& DepthP+P (Ours) & \ding{51} & 0.152 & 1.322 & 6.185 & 0.239 & 0.781 & 0.920 & 0.970 \tabularnewline 
\hline\hline 
\parbox[t]{2mm}{\multirow{10}{*}{\rotatebox[origin=c]{90}{\small{Improved GT \cite{Uhrig2017THREEDV}}}}} 
& Zhou \etal \cite{Zhou2017CVPR} & \ding{55} & 0.176 & 1.532 & 6.129 & 0.244 & 0.758 & 0.921 & 0.971 \tabularnewline
& Mahjourian \etal \cite{Mahjourian2018CVPR} & \ding{55} & 0.134 & 0.983 & 5.501 & 0.203 & 0.827 & 0.944 & 0.981 \tabularnewline
& Yin \etal \cite{Yin2018CVPR} & \ding{55} & 0.132 & 0.994 & 5.240 & 0.193 & 0.833 & 0.953 & 0.985 \tabularnewline
& Wang \etal \cite{Wang2018CVPR} & \ding{55} & 0.126 & 0.866 & 4.932 & 0.185 & 0.851 & 0.958 & 0.986 \tabularnewline
& Ranjan \etal \cite{Ranjan2019CVPR} & \ding{55} & 0.123 & 0.881 & 4.834 & 0.181 & 0.860 & 0.959 & 0.985 \tabularnewline
& Luo \etal \cite{Luo2019PAMI} & \ding{55} & 0.120 & 0.789 & 4.755 & 0.177 & 0.856 & 0.961 & 0.987 \tabularnewline
& Godard \etal \cite{Godard2019ICCV} & \ding{55} & \underline{0.085} & 0.468 & \underline{3.672} & \underline{0.128} & \underline{0.921} & \underline{0.985} & \underline{0.995} \tabularnewline
& Safadoust \etal \cite{Safa2021ThreeDV} & \ding{55} & \underline{0.085} & \underline{0.458} & 3.779 & 0.131 & 0.919 & \underline{0.985} & \textbf{0.996} \tabularnewline
& Guizilini \etal \cite{Guizilini2020CVPR} & \ding{55} & \textbf{0.078} & \textbf{0.420} & \textbf{3.485} & \textbf{0.121} & \textbf{0.931} & \textbf{0.986} & \textbf{0.996} \tabularnewline
& DepthP+P (Ours) & \ding{51} & 0.134 & 1.042 & 5.566 & 0.199 & 0.820 & 0.946 & 0.983
\end{tabular}
\end{adjustbox}
    \vspace{2pt}
    \caption{\textbf{Quantitative Results for Monocular Training on KITTI.} This table compares our proposed approach, \textbf{DepthP+P}, to previous approaches on the KITTI dataset that were trained only with monocular supervision. The \textbf{scale} column specifies whether the method can estimate depth in metric scale.
     We provide results with the original and improved ground truth. We show the results for the input resolution $640 \times 192$. The best method in each column is shown in bold and the second best is underlined.}
    \label{tab:kitti_main}
\end{table*}

%% file: tables/stereo.tex
\begin{table*}[t]
    \centering
    \begin{adjustbox}{max width=\textwidth}
    \begin{tabular}{l | l | c c c c | c c c }
    \toprule
 & {\multirow{2}{*}{\bf Method}}  & \multicolumn{4}{c}{Lower Better} & \multicolumn{3}{c}{Higher Better} \tabularnewline 
& & Abs Rel & Sq Rel & RMSE & RMSE$_{log}$ & $\delta < 1.25$ & $\delta < 1.25^2$ & $\delta < 1.25^3$\tabularnewline \hline
\parbox[t]{2mm}{\multirow{6}{*}{\rotatebox[origin=c]{90}{\small{Original GT}}}}
& Li \etal \cite{Li2018ICRA}  & 0.183 & 1.730 & 6.570 & 0.268 & - & - & - \tabularnewline 
& Zhan \etal \cite{Zhan2018CVPR} &  0.135 & 1.132 & 5.585 & 0.229 & 0.820 & 0.933 & \underline{0.971} \tabularnewline
& Luo \etal \cite{Luo2019PAMI} & 0.128 & 0.935 & 5.011 & 0.209 & 0.831 & 0.945 & \textbf{0.979} \tabularnewline
& Godard \etal \cite{Godard2019ICCV} & \textbf{0.106} & \textbf{0.818} & \textbf{4.750} & \textbf{0.196} & \textbf{0.874} & \textbf{0.957} & \textbf{0.979} \tabularnewline
& DepthP+P (ResNet18) & \underline{0.110} & 0.907 & 4.888 & 0.199 & 0.867 & \underline{0.954} & \textbf{0.979} \tabularnewline
& DepthP+P (ResNet50) & \textbf{0.106} & \underline{0.900} & \underline{4.828} & \underline{0.198} & \underline{0.871} & \underline{0.954} & \textbf{0.979} \tabularnewline
 \hline\hline 
\parbox[t]{2mm}{\multirow{5}{*}{\rotatebox[origin=c]{90}{\small{Improved GT}}}} 
& Zhan \etal \cite{Zhan2018CVPR} &  0.130 & 1.520 & 5.184 & 0.205 & 0.859 & 0.955 & 0.981 \tabularnewline
& Luo \etal \cite{Luo2019PAMI} & 0.123 & 0.754 & 4.453 & 0.172 & 0.863 & 0.964 & 0.989 \tabularnewline
& Godard \etal \cite{Godard2019ICCV} & \textbf{0.080} & \textbf{0.466} & \textbf{3.681} & \textbf{0.127} & \textbf{0.926} & \textbf{0.985} & \textbf{0.995} \tabularnewline
& DepthP+P (ResNet18) & 0.088 & 0.572 & 3.905 & 0.138 & 0.911 & 0.981 & \underline{0.994} \tabularnewline
& DepthP+P (ResNet50) & \underline{0.084} & \underline{0.543} & \underline{3.784} & \underline{0.134} & \underline{0.916} & \underline{0.982} & \textbf{0.995} \tabularnewline
\end{tabular}
\end{adjustbox}
    \vspace{2pt}
    \caption{\textbf{Quantitative Results on KITTI with Additional Stereo Supervision.} We compare DepthP+P to previous approaches that use additional stereo supervision on KITTI. 
 Stereo supervision significantly improves the results of DepthP+P model. By using ResNet50, our model performs on par with Monodepth2~\cite{Godard2019ICCV}.}
    \label{tab:stereo}
\end{table*}

%% file: figures/noground.tex
\begin{figure*}[t]
    \includegraphics[width=\linewidth]{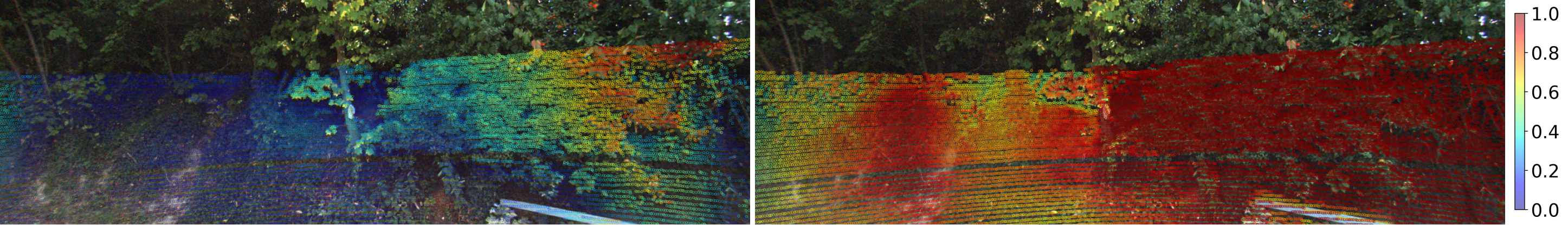}
    \caption{\textbf{Qualitative Comparison.} We compare the absolute relative error of our depth estimation method (\textbf{left}) with DNet \cite{Xue2020IROS} (\textbf{right}) on an image from the KITTI dataset without a ground plane. The colorbar on the right shows the values of the absolute relative error metric. We cap the max error at the value of 1.0 for visualization. We can see that \cite{Xue2020IROS} completely fails to estimate the metric depth due to the wrong scale recovery because there is no ground plane in the image, while our model does not have this issue and can still perform well. The absolute relative error for \cite{Xue2020IROS} is 1.178 while it is 0.252 for our method.   
    }
    \label{fig:noground}
\end{figure*}

%% file: sec/conclusion.tex
\section{Conclusion and Future Work}
\label{sec:conclusion}
In this paper, we presented a new approach to self-supervised monocular depth estimation following the traditional planar parallax paradigm. We showed that our approach is able to produce metrically accurate depth estimates by using a known camera height. Unlike previous methods that rely on estimating the full rigid-body motion of the camera, our method only needs to estimate the camera translation. We discussed the advantage of our method compared to the other scale-aware depth prediction methods. We see our approach as a first step to unlocking the potential of the plane and parallax for efficient and metric-accurate depth estimation. An exciting future direction can focus on detecting moving foreground objects by checking the violations in the plane and parallax constraints~\cite{Irani1996ECCV}.


%% file: supp/sec/summary.tex
In \secref{supp_sec:derivation} of this supplementary document, we provide the derivation of the residual parallax (Equation 2 in the main paper)~\cite{Irani1996ECCV}. In \secref{supp_sec:normal}, we investigate the effect of using a more accurate estimation of the normal vector of the road. Finally, in \secref{supp_sec:qual}, we provide additional qualitative results of our model.

%% file: supp/sec/derivation.tex
\section{Derivation}
\label{supp_sec:derivation}
Let $\bC$ and $\bC'$ be the camera centers of the target view $\bI_t$ and the source view $\bI_s$, respectively. Also, let $\bx=[\bX, \bY, \bZ]^T$ and $\bx'=[\bX', \bY', \bZ']^T$ be the coordinates of a 3D point with respect to $\bC$ and $\bC'$ respectively. We can relate $\bx$ and $\bx'$ as follows:
\begin{equation}
    \label{supp_eq:rt}
    \bx = \bR \bx'+\bt
\end{equation}
where $\bR$ is the rotation and $\bt = [\bt_x, \bt_y, \bt_z]^T$ is the translation between $\bC$ and $\bC'$.
According to the Fig 2 (right), we can calculate $\bh$, the perpendicular distance of the 3D point to the plane $\bPi$ as:
\begin{equation}
    \label{supp_eq:height}
    \bh = \bd_c - \bN^T\bx'
\end{equation}
Note that $\bd_c$ is the height of $\bC'$, the camera of the source view $\bI_s$. However, $\bh$ is invariant with respect to the cameras. We can rewrite the above equation as:
\begin{equation}
    1 = \frac{\bh + \bN^T\bx'}{\bd_c}  
\end{equation}
Therefore, by substituting this into \eqnref{supp_eq:rt}, we have:
\begin{equation}
    \label{supp_eq:onesubed}
     \bx = \bR \bx'+\bt \left( \frac{\bh + \bN^T\bx'}{\bd_c}  \right) = \left( \bR + \frac{\bt\bN^T}{\bd_c} \right)\bx' + \frac{\bh}{\bd_c}\bt
\end{equation}
%
Let $\bK$ and $\bK'$ denote the camera intrinsic for the target view and the source view. Then $\bp=[x,y,1]^T=\frac{1}{\bZ}\bK \bx$ and $\bp'=[x',y',1]=\frac{1}{\bZ'}\bK' \bx'$ represent the pixel coordinates of our 3D point in the target and source view, respectively. Therefore \eqnref{supp_eq:onesubed} can be written as:
\begin{equation}
     \bZ \bK^{-1}\bp = \left( \bR + \frac{\bt\bN^T}{\bd_c} \right)\bZ' \bK'^{-1}\bp' + \frac{\bh}{\bd_c} \bt
\end{equation}
By multiplying both sides by $\frac{1}{\bZ'}\bK$ we obtain:

%
\begin{equation}
     \frac{\bZ}{\bZ'} \bp =  \underbrace{\bK \left( \bR + \frac{\bt\bN^T}{\bd_c} \right) \bK'^{-1}}_{\bH} \bp' + \frac{\bh}{\bZ' \bd_c} \bK \bt
\end{equation}

\begin{equation}
    \label{supp_eq:Hfirst}
     \frac{\bZ}{\bZ'} \bp = \bH\bp' + \frac{\bh}{\bZ' \bd_c} \bK \bt
\end{equation}
where $\bH = \bK \left( \bR + \frac{\bt\bN^T}{\bd_c} \right) \bK'^{-1}$ is the $3\times 3$ homography matrix associated with the plane $\bPi$ between the source and the target view.
Note that in \eqnref{supp_eq:Hfirst}, the third component of the vector in two sides of the equation should be equal. Therefore:

\begin{equation}
    \label{}
     \frac{\bZ}{\bZ'}= \bH_3\bp' + \frac{\bh}{\bZ' \bd_c} \bK_3 \bt
\end{equation}
where $\bH_3$ and $\bK_3$ are the third row of $\bH$ and $\bK$. Since $\bK_3 \bt = \bt_z$, we therefore have:
\begin{equation}
    \label{supp_eq:thirdcomponent}
     \frac{\bZ}{\bZ'}= \bH_3\bp' + \frac{\bh}{\bZ' \bd_c} \bt_z
\end{equation}
By dividing each side of \eqnref{supp_eq:Hfirst} by each side of \eqnref{supp_eq:thirdcomponent} we obtain:
\begin{equation}
     \bp = \frac{\bH\bp' + \frac{\bh}{\bZ' \bd_c} \bK \bt} {\bH_3\bp' + \frac{\bh}{\bZ' \bd_c} \bt_z}
\end{equation}
Adding and subtracting $\frac{\bH\bp'}{\bH_3\bp'}$ to the right-hand side yields:
\begin{flalign}
\bp & = \frac{\bH\bp'}{\bH_3\bp'} - \frac{\bH\bp'}{\bH_3\bp'} +  \frac{\bH\bp' + \frac{\bh}{\bZ' \bd_c} \bK \bt} {\bH_3\bp' + \frac{\bh}{\bZ' \bd_c} \bt_z} \\
     & = \frac{\bH\bp'}{\bH_3\bp'} - \frac{\bH\bp' \left( \bH_3\bp' + \frac{\bh}{\bZ' \bd_c} \bt_z \right) }{\bH_3\bp' \left( \bH_3\bp' + \frac{\bh}{\bZ' \bd_c} \bt_z \right) } + \frac{\bH_3\bp' \left( \bH\bp' + \frac{\bh}{\bZ' \bd_c} \bK \bt \right)}{\bH_3\bp' \left( \bH_3\bp' + \frac{\bh}{\bZ' \bd_c} \bt_z \right) }\\
     \label{supp_eq:long}& = \frac{\bH\bp'}{\bH_3\bp'} - \frac{\bH\bp' \left( \frac{\bh}{\bZ' \bd_c} \bt_z \right) }{\bH_3\bp' \left( \bH_3\bp' + \frac{\bh}{\bZ' \bd_c} \bt_z \right) } + \frac{\bH_3\bp' \left(\frac{\bh}{\bZ' \bd_c} \bK \bt \right)}{\bH_3\bp' \left( \bH_3\bp' + \frac{\bh}{\bZ' \bd_c} \bt_z \right) }
\end{flalign}
%
%
%
Substituting \eqnref{supp_eq:thirdcomponent} into the denominators of \eqnref{supp_eq:long} results in:
\begin{flalign}
    \bp & = \frac{\bH\bp'}{\bH_3\bp'} - \frac{\bH\bp' \left( \frac{\bh}{\bZ' \bd_c} \bt_z \right) }{\bH_3\bp' \left( \frac{\bZ}{\bZ'} \right) } + \frac{\bH_3\bp' \left(\frac{\bh}{\bZ' \bd_c} \bK \bt \right)}{\bH_3\bp' \left( \frac{\bZ}{\bZ'} \right) }\\
    \label{supp_eq:to_replace_H}& = \frac{\bH\bp'}{\bH_3\bp'} - \frac{\bh \bt_z}{\bZ \bd_c}\frac{\bH\bp'}{\bH_3\bp'} + \frac{\bh}{\bZ \bd_c}\bK \bt 
\end{flalign}
%
%
%
The point $\bp_w = [x_w, y_w, 1]^T = \frac{\bH\bp'}{\bH_3\bp'}$ is the point $\bp'$ transformed by the homography matrix $\bH$. Thus, \eqnref{supp_eq:to_replace_H} can be simplified to:
\begin{equation}
    \label{supp_eq:almost_there}
     \bp = \bp_w - \frac{\bh \bt_z}{\bZ \bd_c}\bp_w + \frac{\bh}{\bZ \bd_c}\bK \bt 
\end{equation}
Subtracting $\frac{\bh \bt_z}{\bZ \bd_c}\bp$ from both sides, we obtain:
\begin{equation}
     \bp - \frac{\bh \bt_z}{\bZ \bd_c}\bp = \bp_w - \frac{\bh \bt_z}{\bZ \bd_c}\bp_w + \frac{\bh}{\bZ \bd_c}\bK \bt - \frac{\bh \bt_z}{\bZ \bd_c}\bp
\end{equation}
\begin{equation}
     \bp \left(1- \frac{\bh \bt_z}{\bZ \bd_c}\right) = \bp_w \left(1- \frac{\bh \bt_z}{\bZ \bd_c}\right) + \frac{\bh}{\bZ \bd_c}\bK \bt   - \frac{\bh \bt_z}{\bZ \bd_c}\bp
\end{equation}
By dividing both sides by $1- \frac{\bh \bt_z}{\bZ \bd_c}$ we get:
\begin{equation}
     \bp  = \bp_w  + \frac{\frac{\bh}{\bZ \bd_c}\bK \bt   - \frac{\bh \bt_z}{\bZ \bd_c}\bp}{1- \frac{\bh \bt_z}{\bZ \bd_c}} 
\end{equation}
By rearranging the terms and defining $\bgamma = \frac{\bh}{\bZ}$ we obtain:
\begin{flalign}
    \bp_w - \bp & = \frac{\frac{\bh \bt_z}{\bZ \bd_c}\bp - \frac{\bh}{\bZ \bd_c}\bK \bt   }{1- \frac{\bh \bt_z}{\bZ \bd_c}} \\
    & = \frac{\frac{\bh \bt_z}{\bZ}\bp - \frac{\bh}{\bZ}\bK \bt   }{\bd_c- \frac{\bh \bt_z}{\bZ}}  \\
    & = \frac{\bgamma\bt_z\bp - \bgamma\bK \bt   }{\bd_c- \bgamma\bt_z} \\
    & = \frac{\bgamma}{\bd_c- \bgamma\bt_z} \left( \bt_z\bp - \bK \bt   \right)
\end{flalign}





%% file: supp/sec/normal.tex
\section{Ablation Study on Normal Vector}
\label{supp_sec:normal}
\input{supp/tables/normal}
In our experiments in the paper, we have assumed that the road plane is horizontal with respect to the camera. In other words, we have assumed that $\bN=[0,1,0]^T$. However, the road planes are not always completely flat and can be titled. For example, one side can be higher than the other. Or consider an uphill where the road is sloping upwards. To analyze this, we perform an experiment where we calculate the normal vector of the road using ground truth depth. Concretely, during training, we back-project the road pixels to 3D using their ground truth depth and fit a plane to the obtained 3D points. We then use the normal vector of the fitted plane as our vector $\bN$ in Equation 4 in the main paper. We report the results in \tabref{supp_tab:normal}. It can be seen that by estimating $\bN$ using this approach, we can achieve much better results. In this experiment, the ground truth values were only used to calculate the normal vector $\bN$. Therefore, we conclude that our method can benefit from a more accurate estimation of the normal vector. A future study can focus on predicting $\bN$ more accurately during training, as was done in \cite{Xue2020IROS} in testing phase, in Planar Parallax framework. 

%% file: supp/tables/normal.tex
\begin{table*}[b]
    \centering
    \begin{adjustbox}{max width=\textwidth}
    \begin{tabular}{l|cccc|ccc}
        \toprule
        Method &  Abs Rel  & Sq Rel & RMSE & RMSE$_{log}$ & $\delta < 1.25$ & $\delta < 1.25^2$ & $\delta < 1.25^3$ \tabularnewline 
        \midrule
         DepthP+P (Fixed Normal)  & 0.152 & 1.322 & 6.185 & 0.239 & \textbf{0.781} & 0.920 & 0.970
        \tabularnewline
        DepthP+P (GT Normal)  & \textbf{0.142} & \textbf{1.127} & \textbf{5.922} & \textbf{0.238} & 0.780 & \textbf{0.921} & \textbf{0.971} \tabularnewline
       \bottomrule
    \end{tabular}
    \end{adjustbox}
    \vspace{3pt}
    \caption{\textbf{Effect of Calculating Normal Vector of Road.}  This table analyzes the effect of accurately predicting the normal vector $\mathbf{N}$ of the road on the performance. Fixed Normal assumes that the ground is perfectly horizontal. GT Normal uses the ground truth depth data, generates the point clouds for the road region, fits a plane and calculates the normal vector. GT Normal performs better than Fixed Normal, highlighting the fact that a more accurate normal vector calculation will improve the performance.}
    \label{supp_tab:normal}
\end{table*}

%% file: supp/sec/qual.tex
\section{Qualitative Results}
\label{supp_sec:qual}
In \figref{supp_fig:qualitative_results}, we provide qualitative results of our models on the KITTI dataset in comparison to Monodepth2~\cite{Godard2019ICCV} and DNet~\cite{Xue2020IROS}. Our stereo model produces sharp outputs and captures the boundaries of the objects very well, and neither of our models suffer from artefacts such as the wrong estimation for the road lane line in the last row.
\input{supp/figures/qual}

%% file: supp/figures/qual.tex
\begin{figure*}[h]
    \def\hspacing{0.002\linewidth}
    \def\vspacing{0.03cm}
    \def\imgw{0.198\linewidth}
    \begin{tabular}{ccccc}
    \hspace{0.016\linewidth}Input Image &  \hspace{0.0015\linewidth} Monodepth2~\cite{Godard2019ICCV} &  \hspace{0.023\linewidth}DNet~\cite{Xue2020IROS} &
    \hspace{0.055\linewidth} Ours-Mono & 
    \hspace{0.035\linewidth} Ours-Stereo\\
    \end{tabular}

    \includegraphics[width=\imgw]{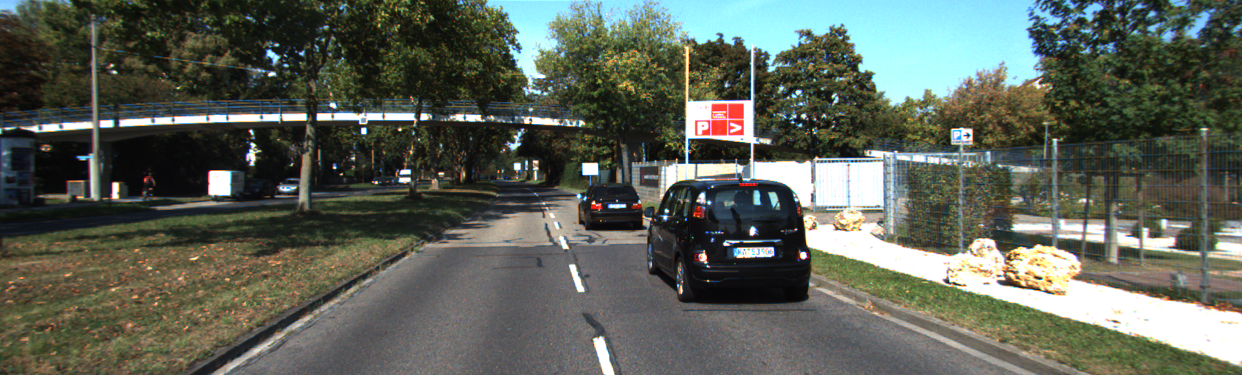}\hspace{\hspacing}%
    \includegraphics[width=\imgw]{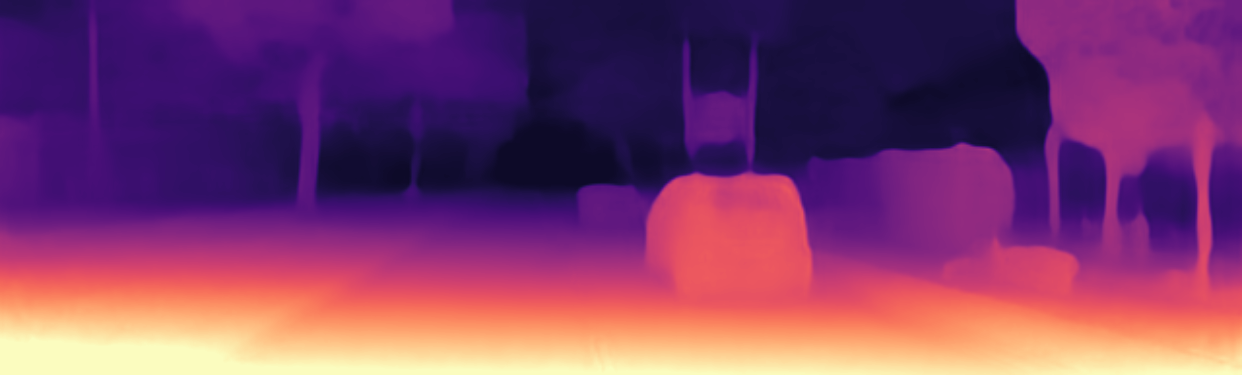}\hspace{\hspacing}%
    \includegraphics[width=\imgw]{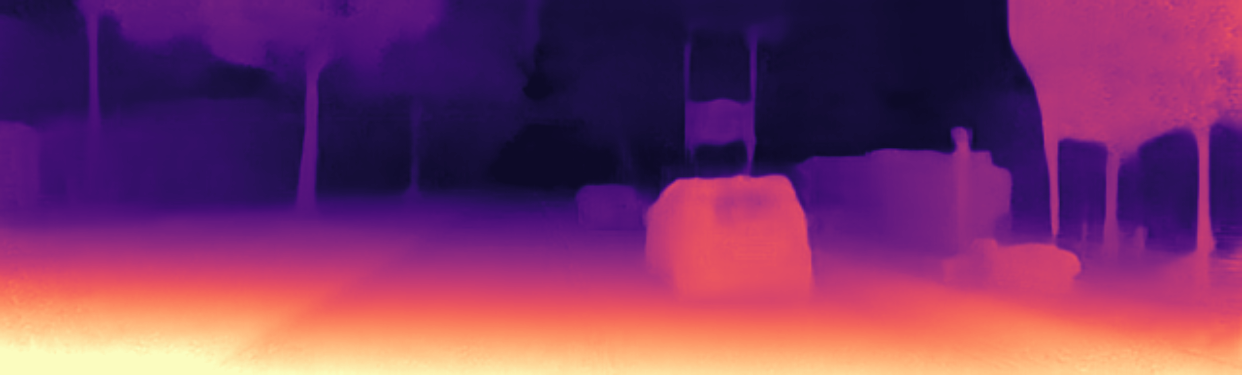}\hspace{\hspacing}%
    \includegraphics[width=\imgw]{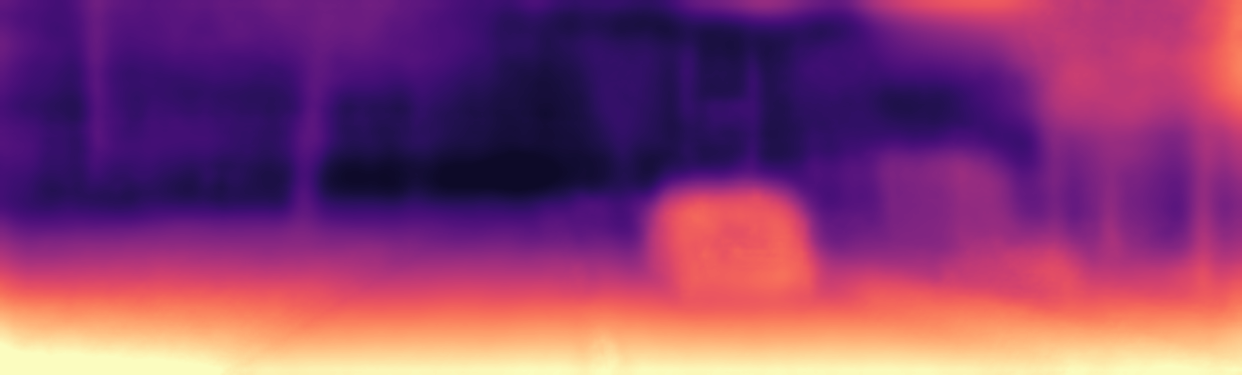}\hspace{\hspacing}%
    \includegraphics[width=\imgw]{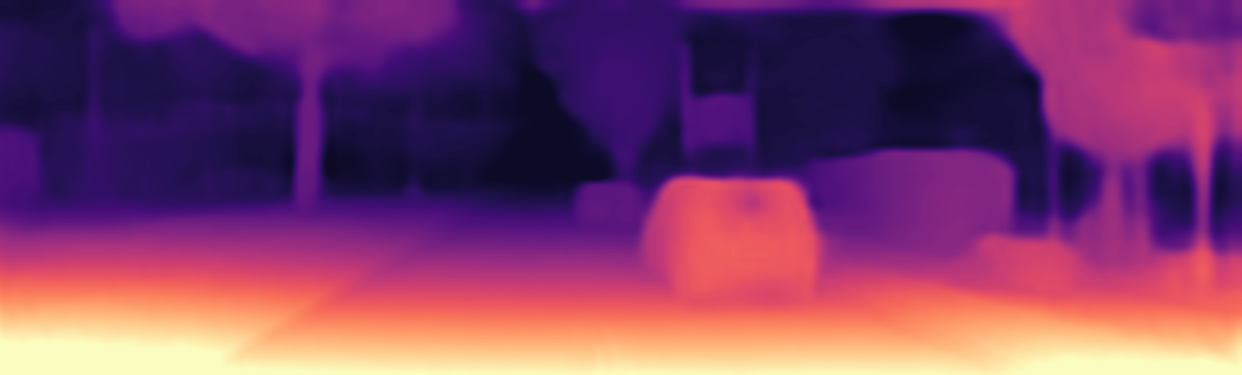}\hspace{\hspacing}%
    \vspace{\vspacing}




\includegraphics[width=\imgw]{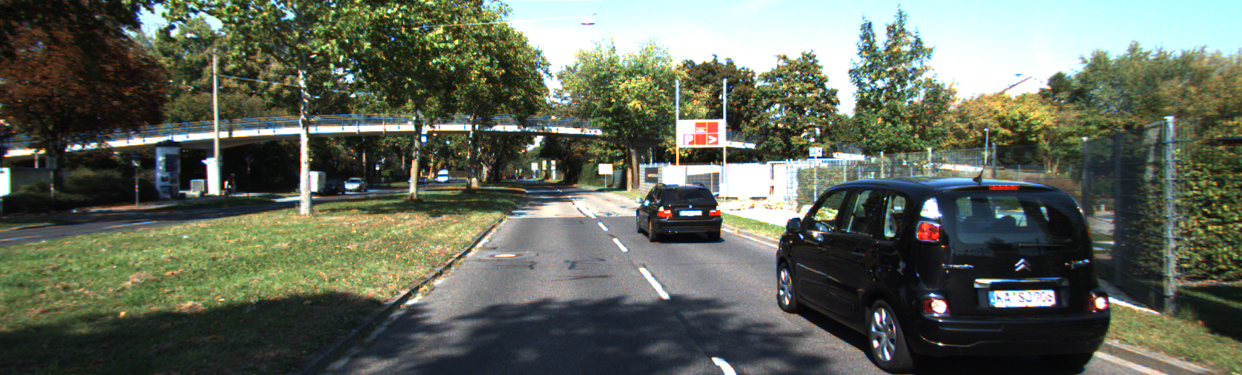}\hspace{\hspacing}%
    \includegraphics[width=\imgw]{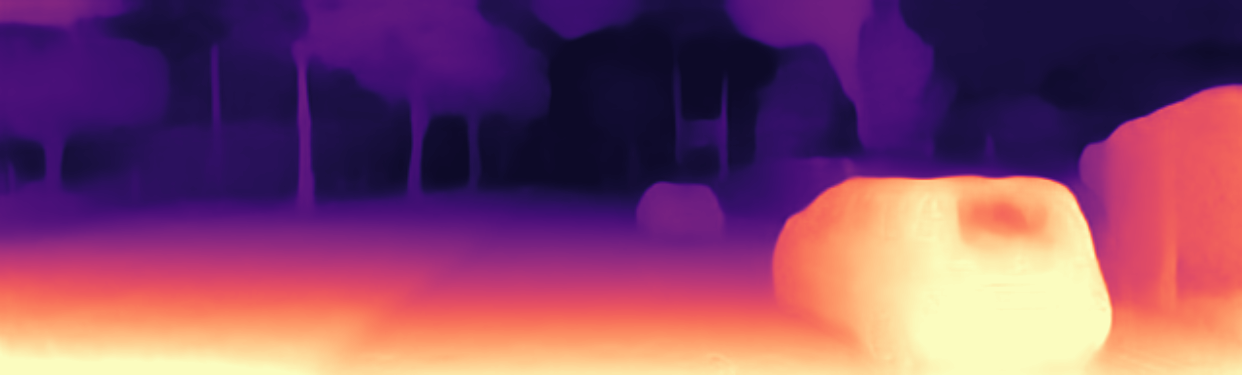}\hspace{\hspacing}%
    \includegraphics[width=\imgw]{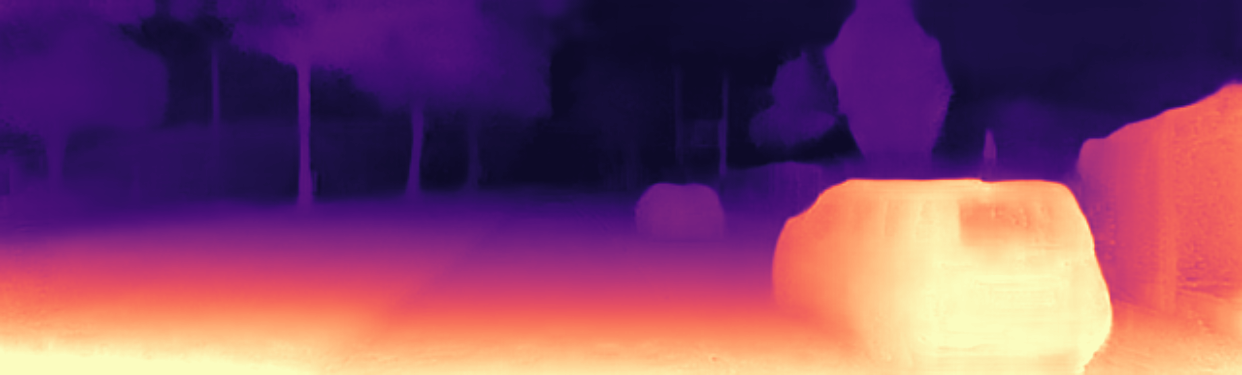}\hspace{\hspacing}%
    \includegraphics[width=\imgw]{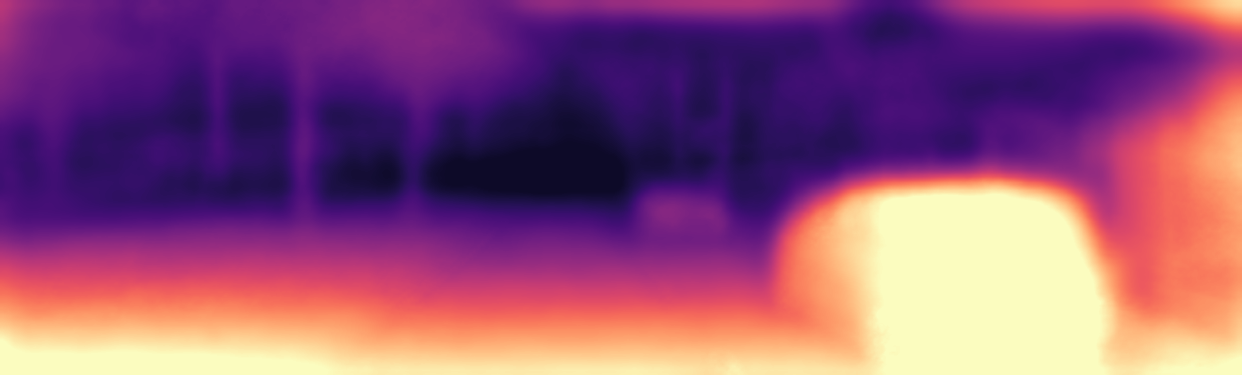}\hspace{\hspacing}%
    \includegraphics[width=\imgw]{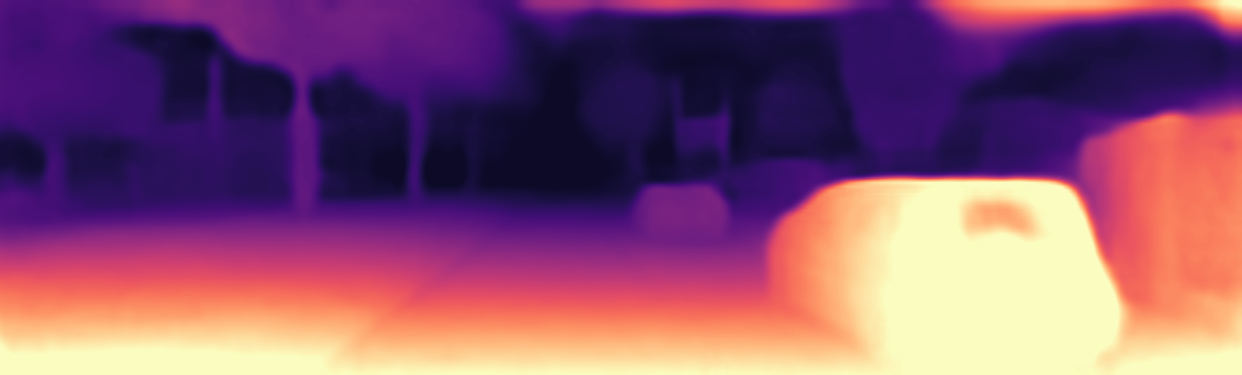}\hspace{\hspacing}%
    \vspace{\vspacing}


\includegraphics[width=\imgw]{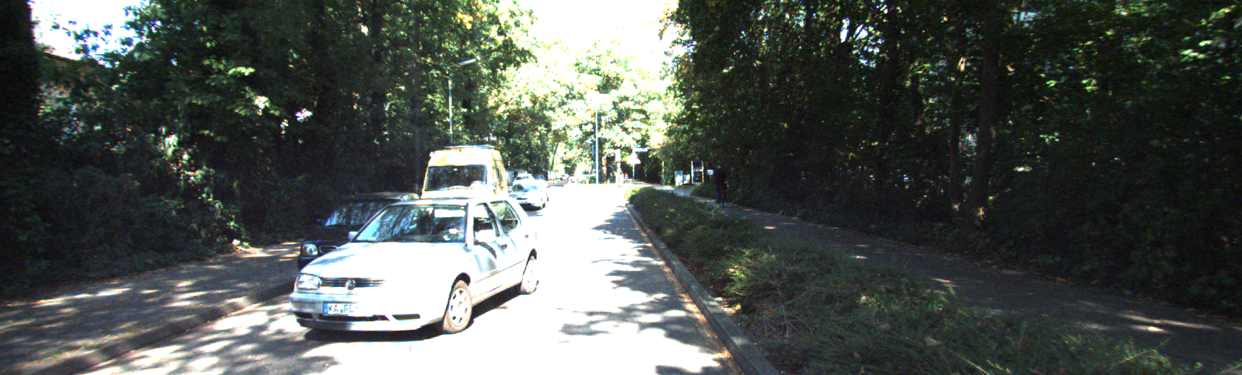}\hspace{\hspacing}%
    \includegraphics[width=\imgw]{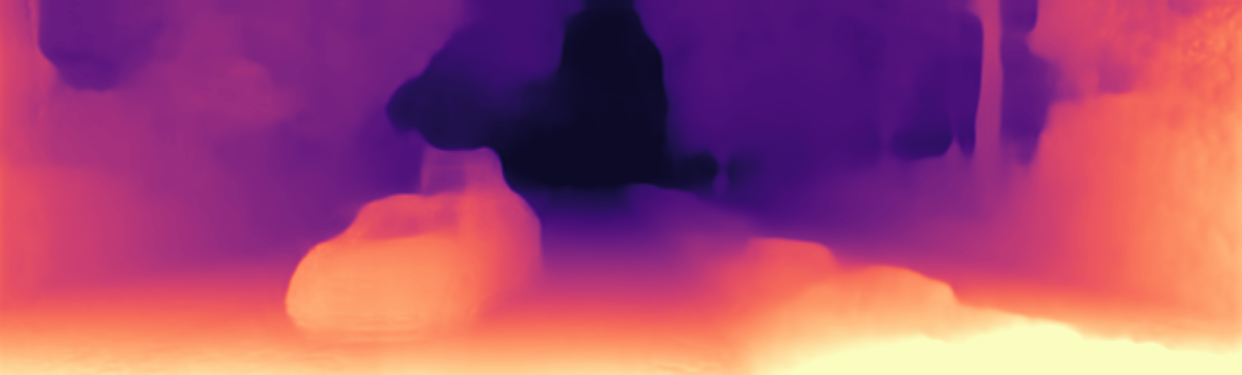}\hspace{\hspacing}%
    \includegraphics[width=\imgw]{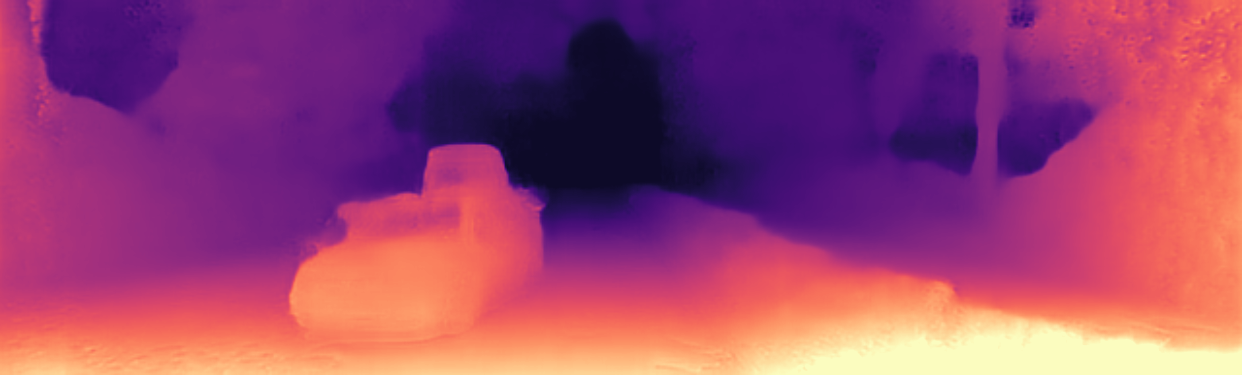}\hspace{\hspacing}%
    \includegraphics[width=\imgw]{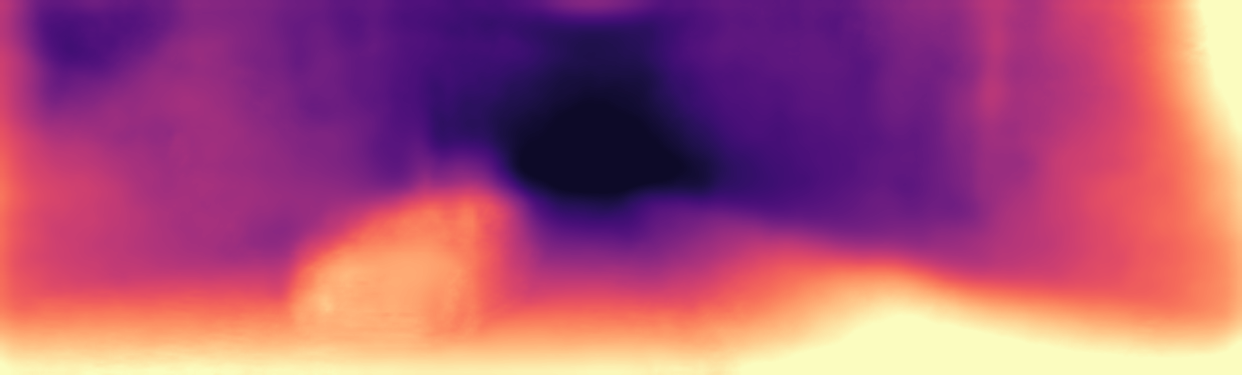}\hspace{\hspacing}%
    \includegraphics[width=\imgw]{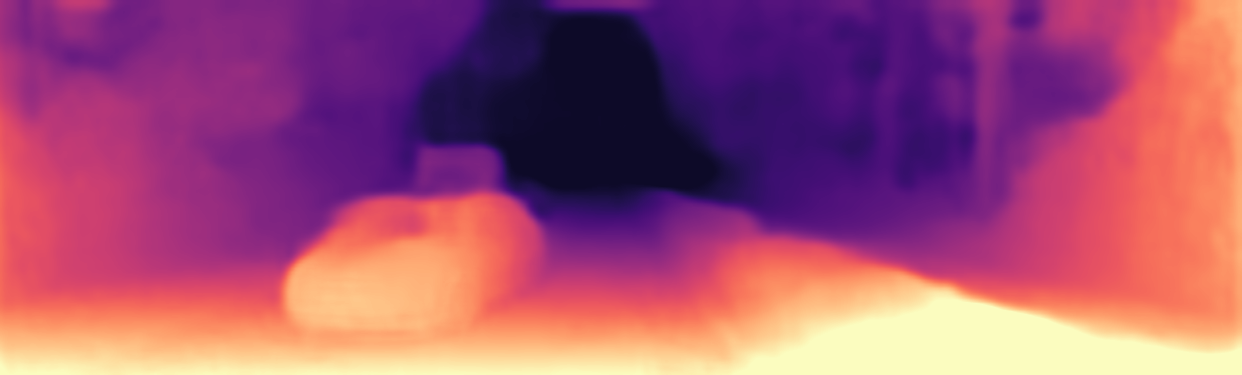}\hspace{\hspacing}%
    \vspace{\vspacing}

\includegraphics[width=\imgw]{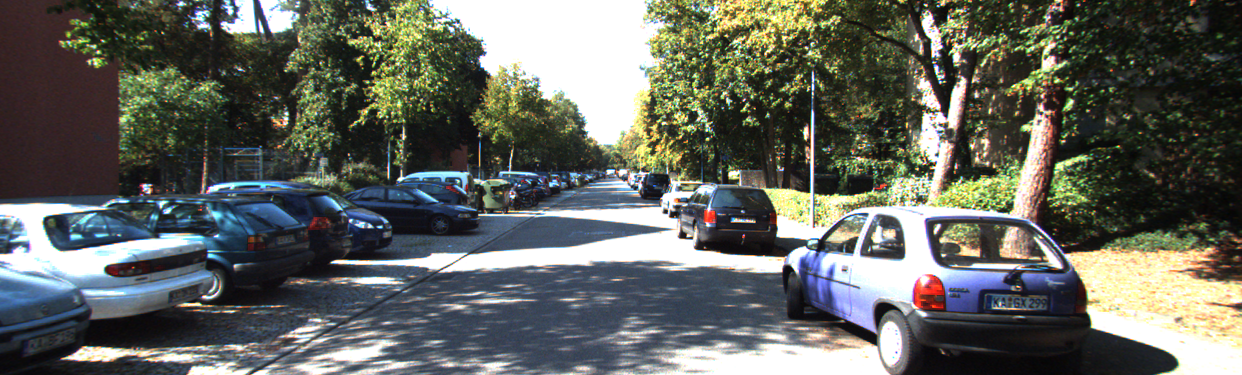}\hspace{\hspacing}%
    \includegraphics[width=\imgw]{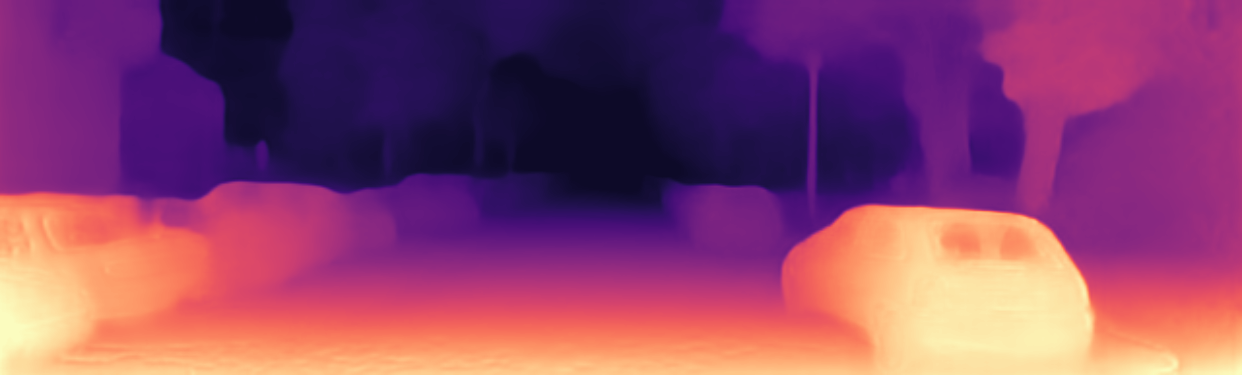}\hspace{\hspacing}%
    \includegraphics[width=\imgw]{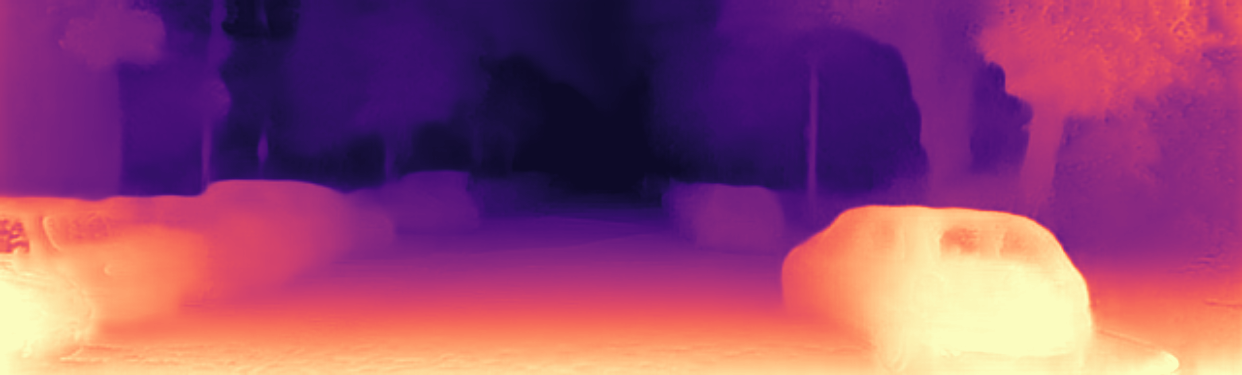}\hspace{\hspacing}%
    \includegraphics[width=\imgw]{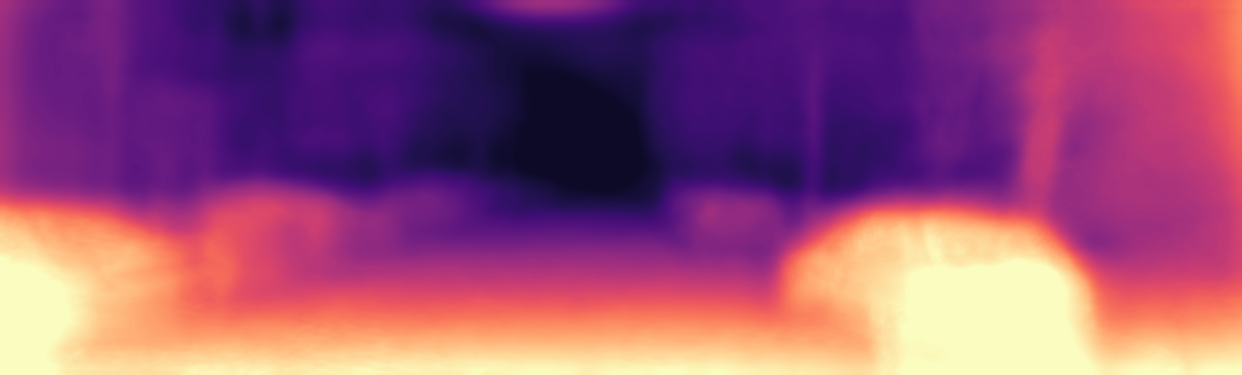}\hspace{\hspacing}%
    \includegraphics[width=\imgw]{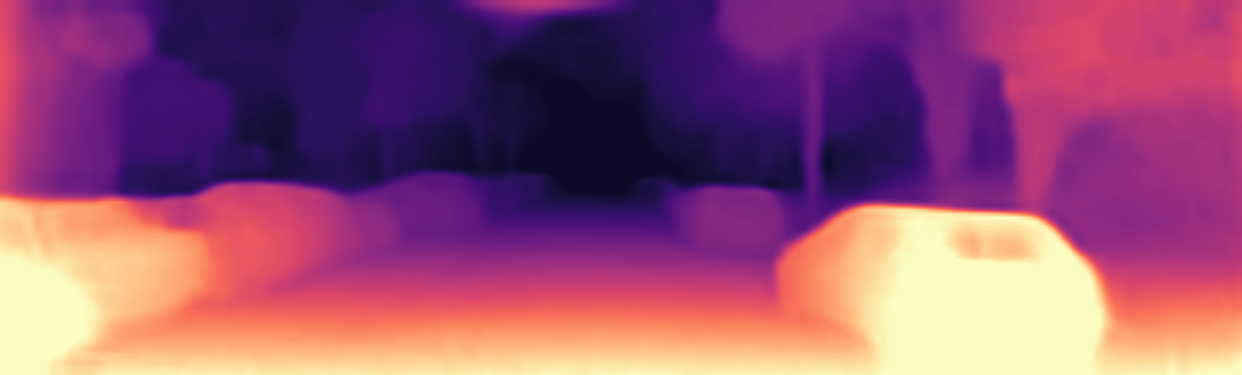}\hspace{\hspacing}%
    \vspace{\vspacing}

\includegraphics[width=\imgw]{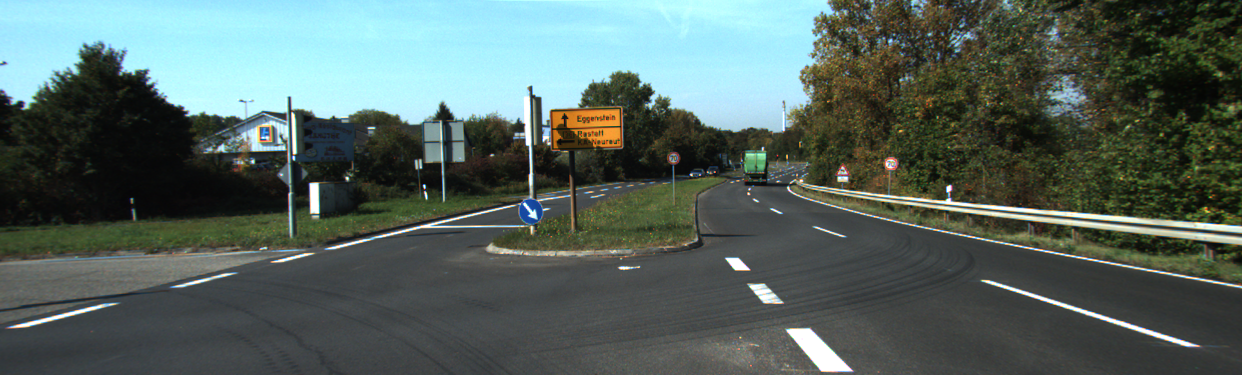}\hspace{\hspacing}%
    \includegraphics[width=\imgw]{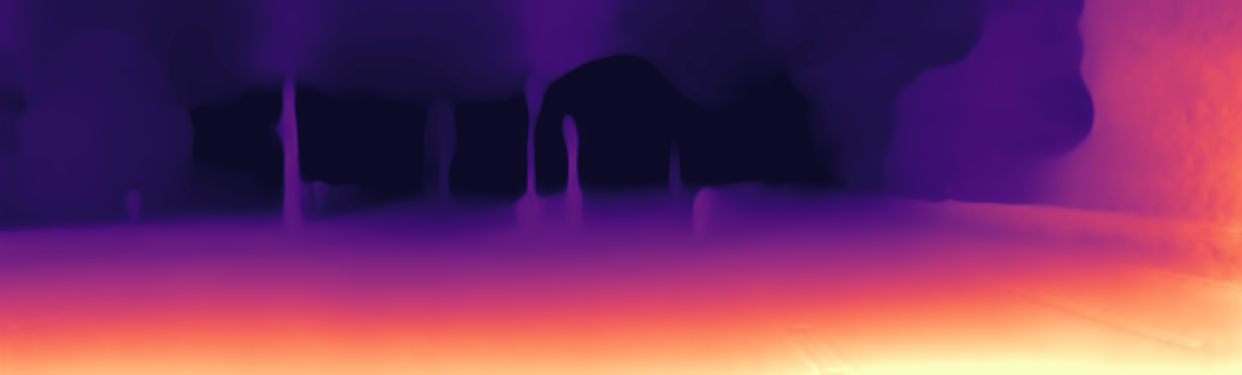}\hspace{\hspacing}%
    \includegraphics[width=\imgw]{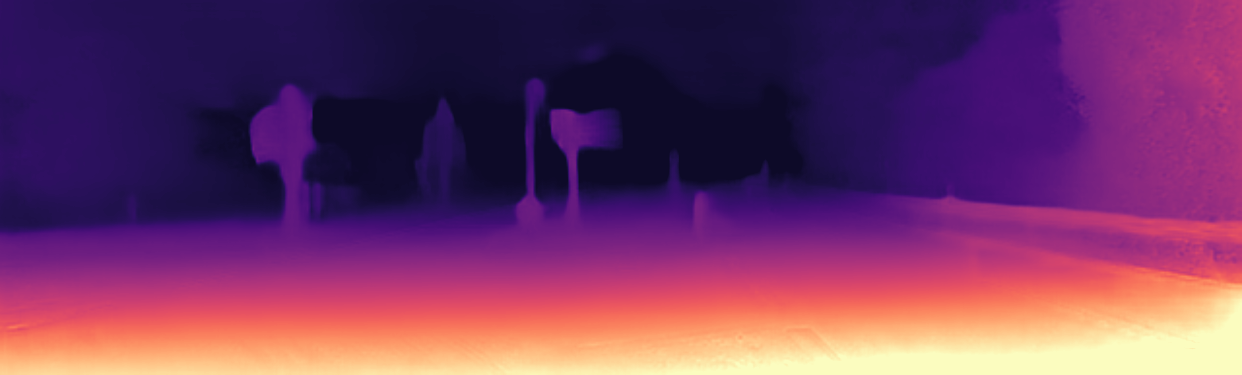}\hspace{\hspacing}%
    \includegraphics[width=\imgw]{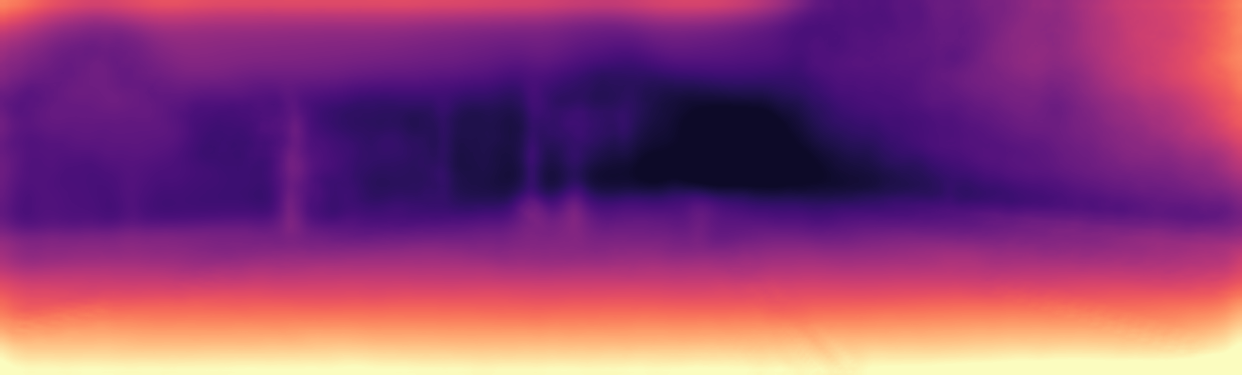}\hspace{\hspacing}%
    \includegraphics[width=\imgw]{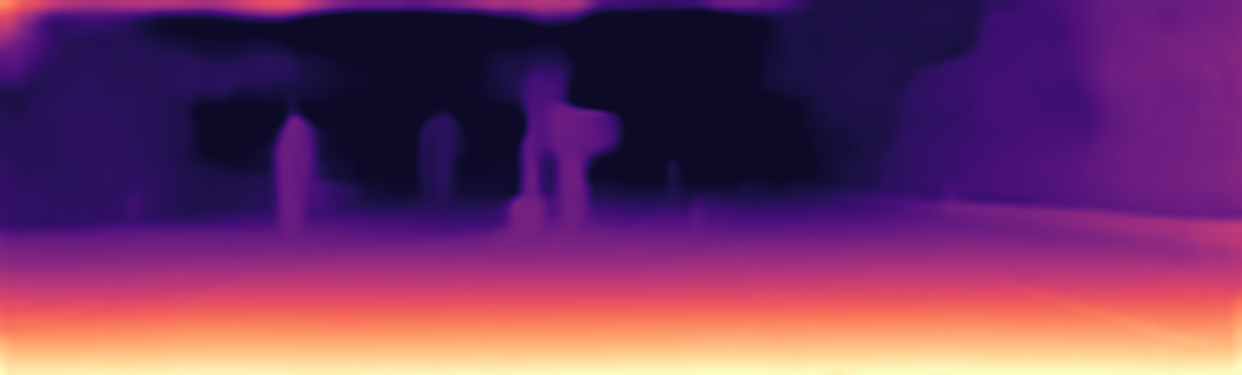}\hspace{\hspacing}%
    \vspace{\vspacing}

\includegraphics[width=\imgw]{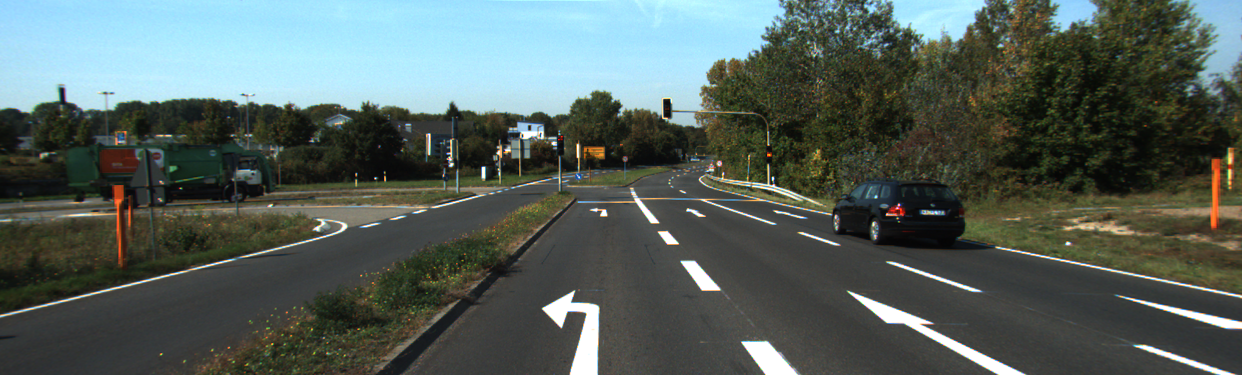}\hspace{\hspacing}%
    \includegraphics[width=\imgw]{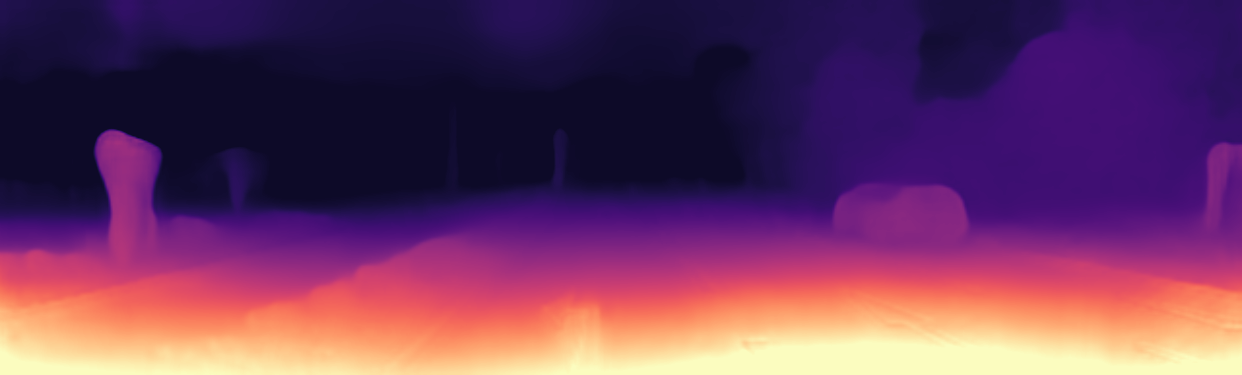}\hspace{\hspacing}%
    \includegraphics[width=\imgw]{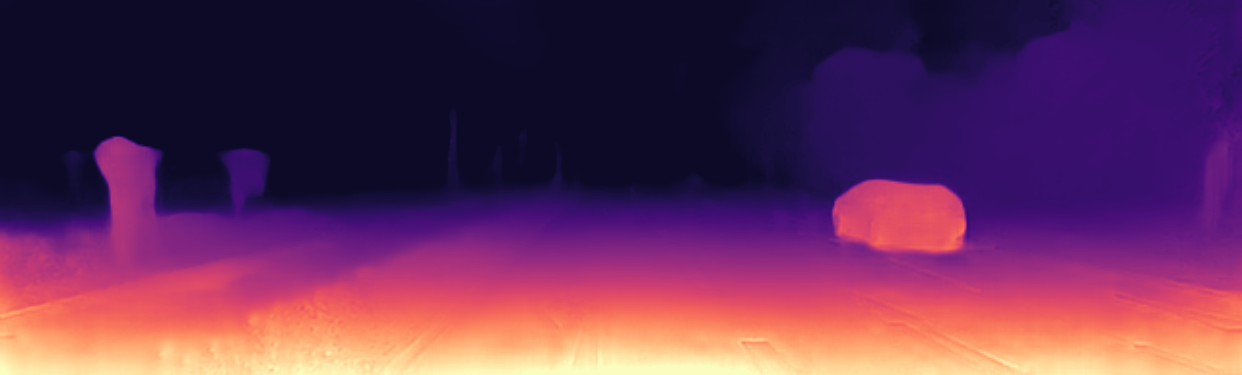}\hspace{\hspacing}%
    \includegraphics[width=\imgw]{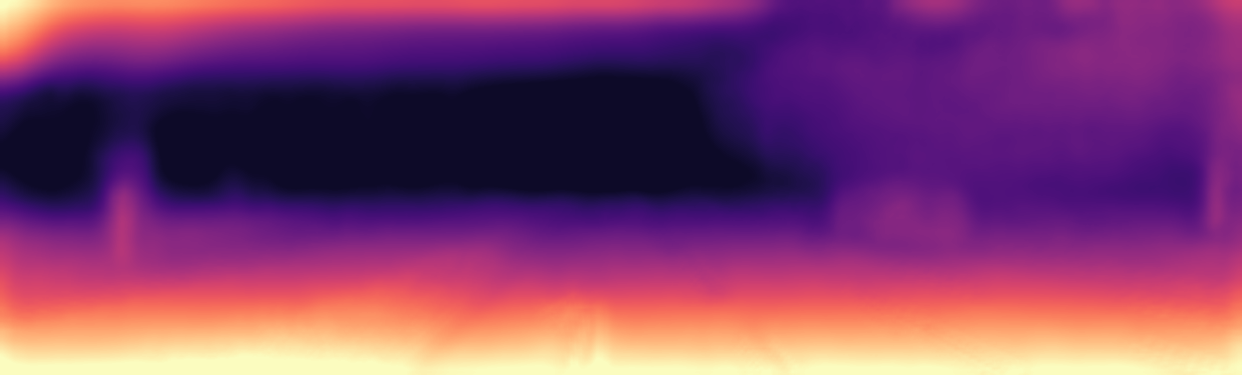}\hspace{\hspacing}%
    \includegraphics[width=\imgw]{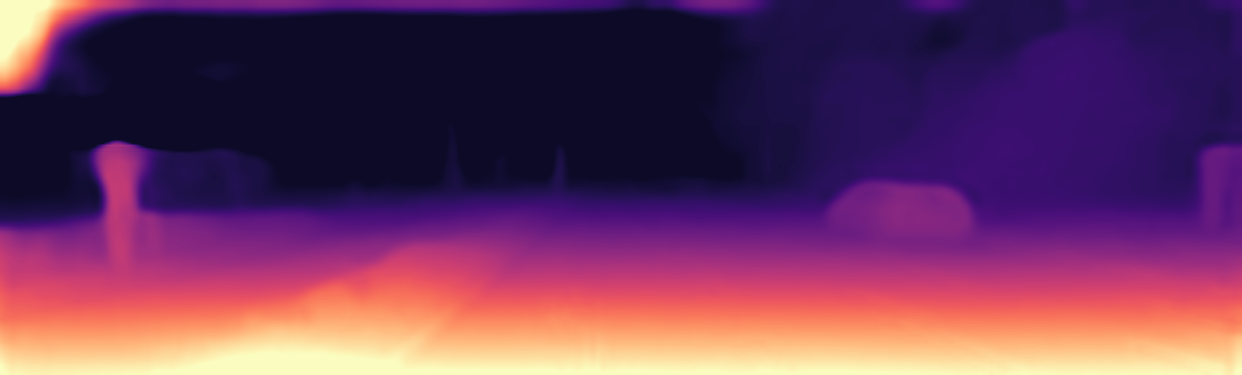}\hspace{\hspacing}%
    \vspace{\vspacing}


\includegraphics[width=\imgw]{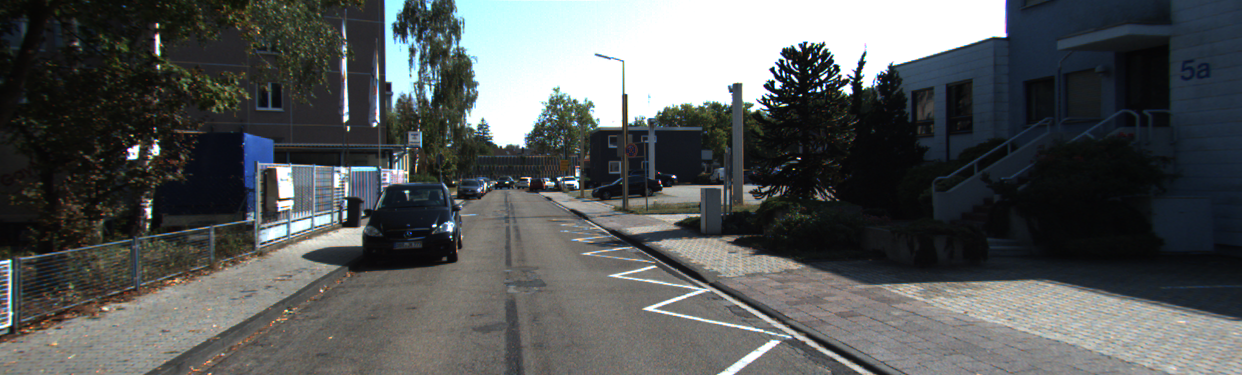}\hspace{\hspacing}%
    \includegraphics[width=\imgw]{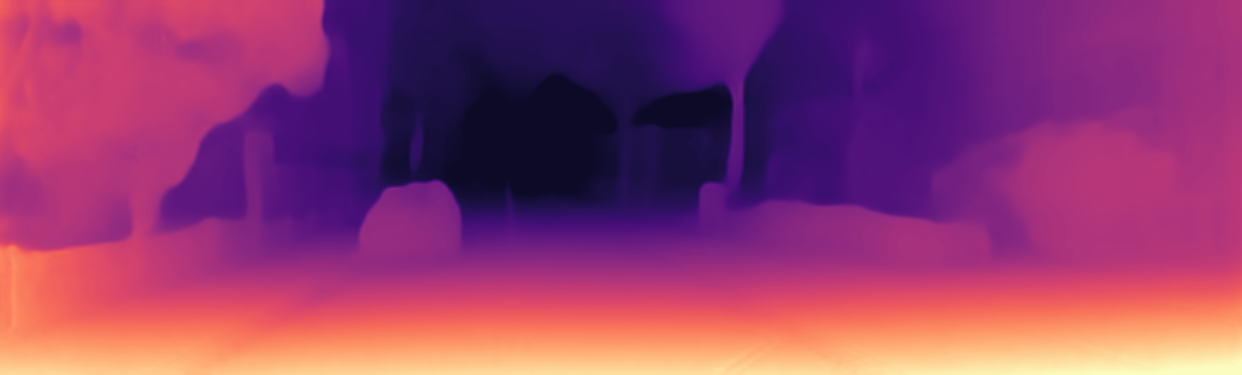}\hspace{\hspacing}%
    \includegraphics[width=\imgw]{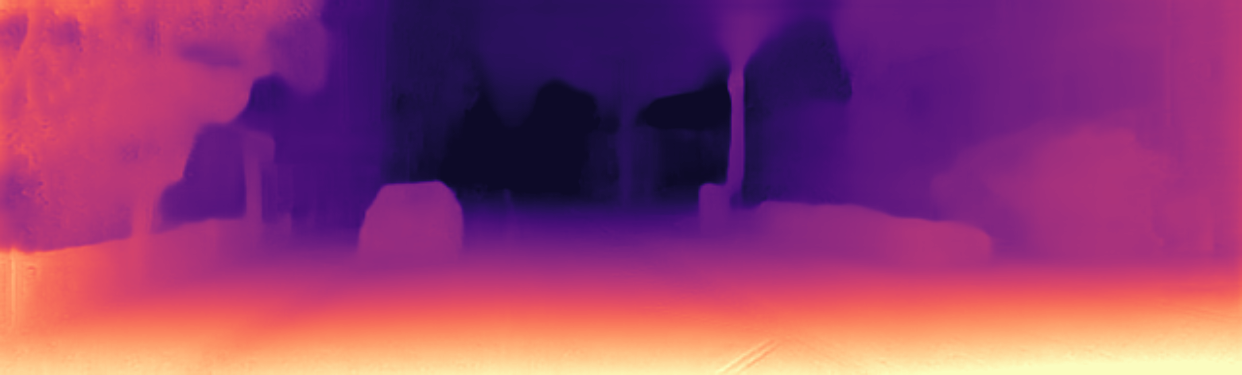}\hspace{\hspacing}%
    \includegraphics[width=\imgw]{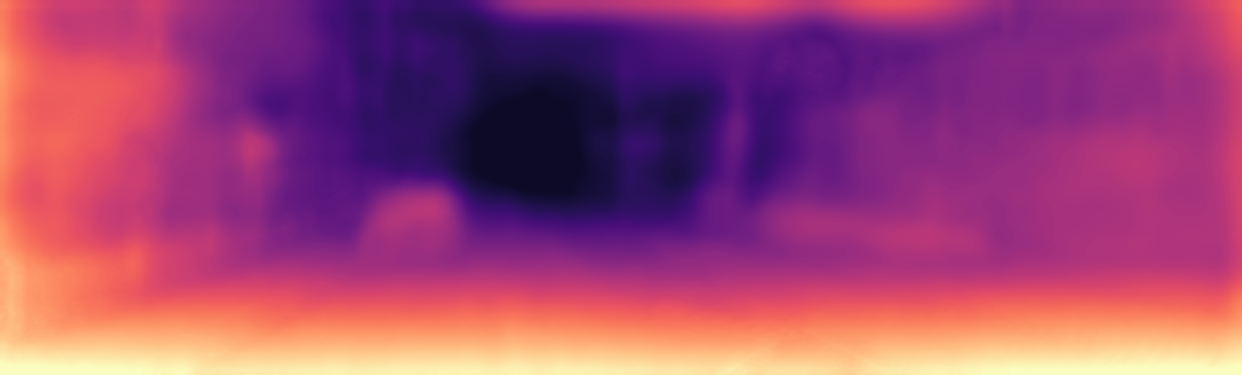}\hspace{\hspacing}%
    \includegraphics[width=\imgw]{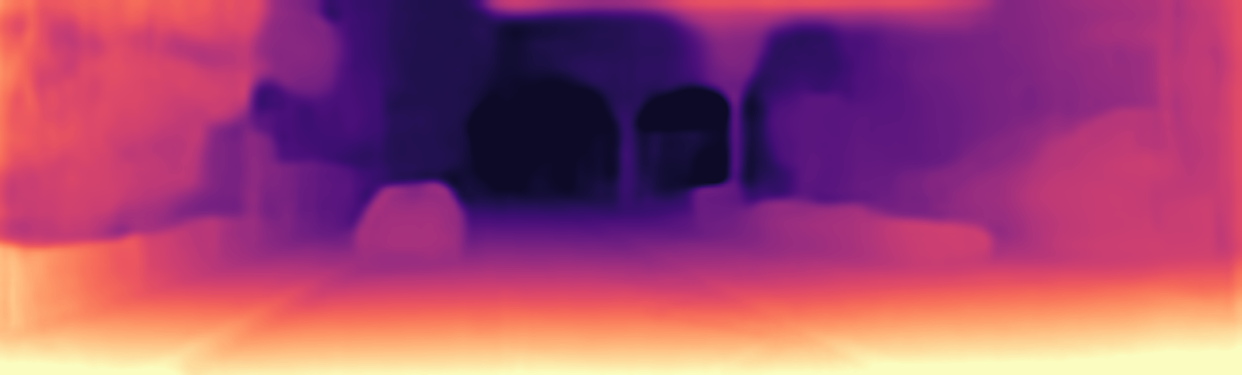}\hspace{\hspacing}%
    \vspace{\vspacing}

\includegraphics[width=\imgw]{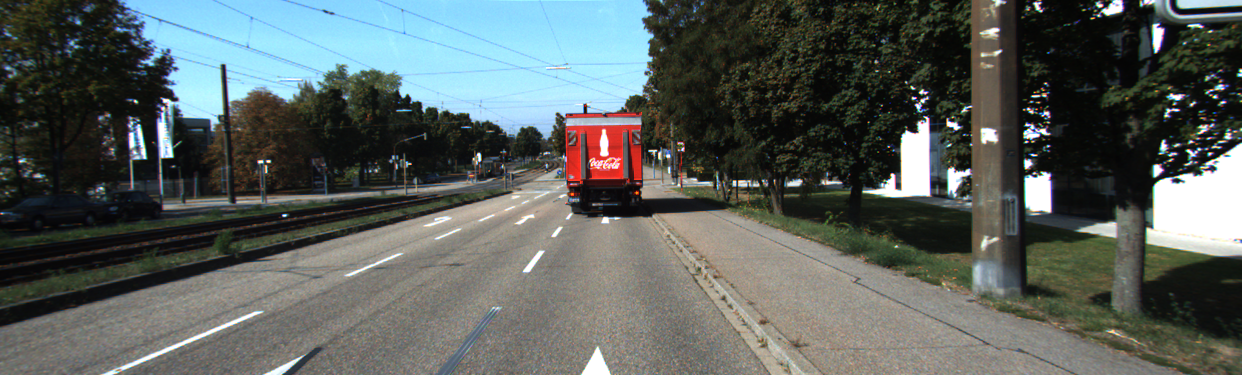}\hspace{\hspacing}%
    \includegraphics[width=\imgw]{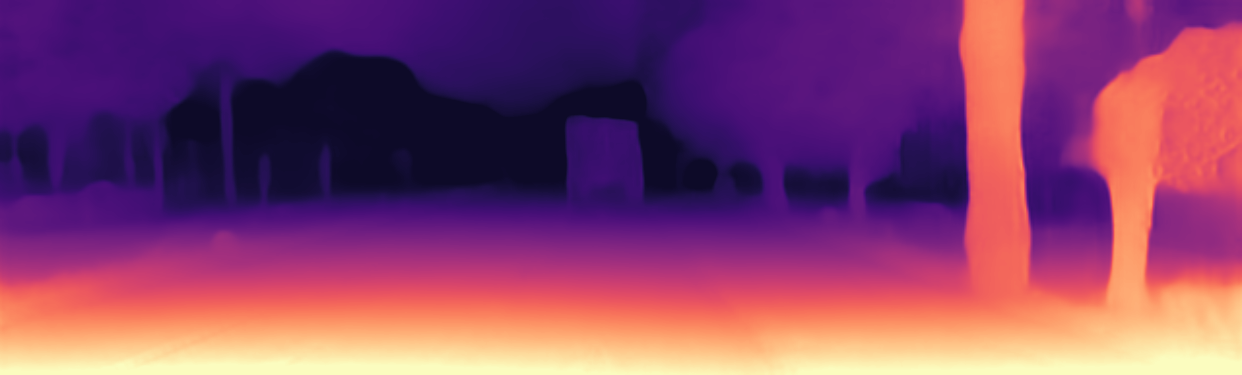}\hspace{\hspacing}%
    \includegraphics[width=\imgw]{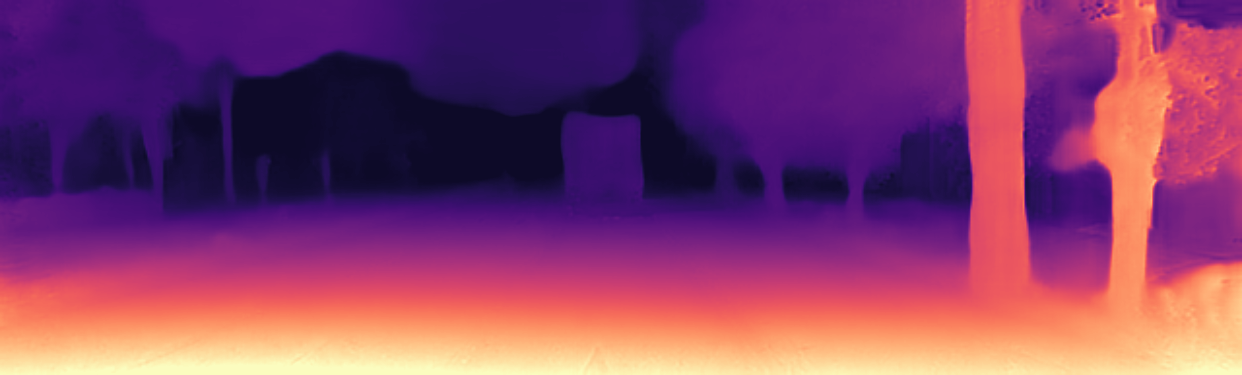}\hspace{\hspacing}%
    \includegraphics[width=\imgw]{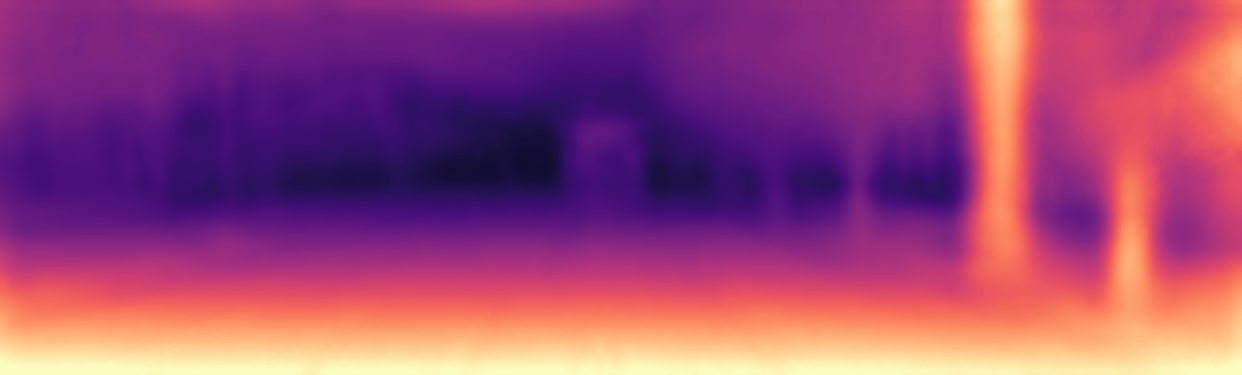}\hspace{\hspacing}%
    \includegraphics[width=\imgw]{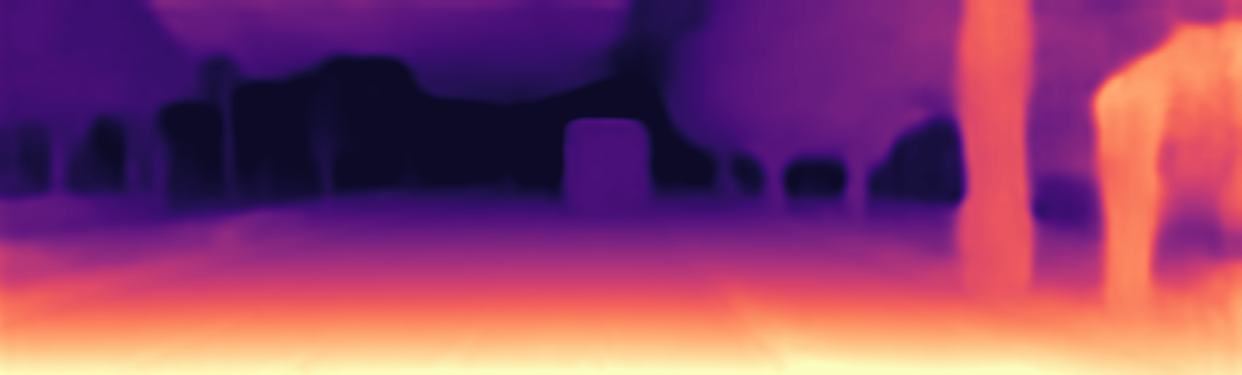}\hspace{\hspacing}%
    \vspace{\vspacing}

\includegraphics[width=\imgw]{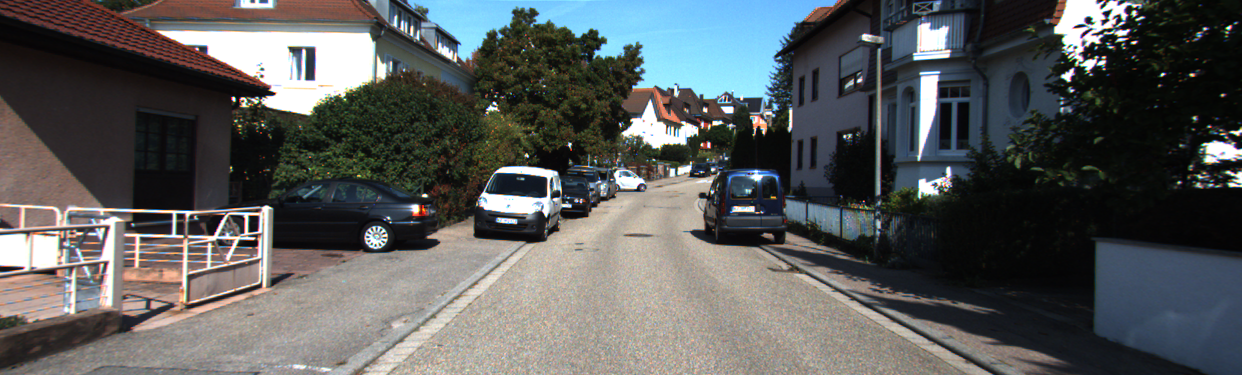}\hspace{\hspacing}%
    \includegraphics[width=\imgw]{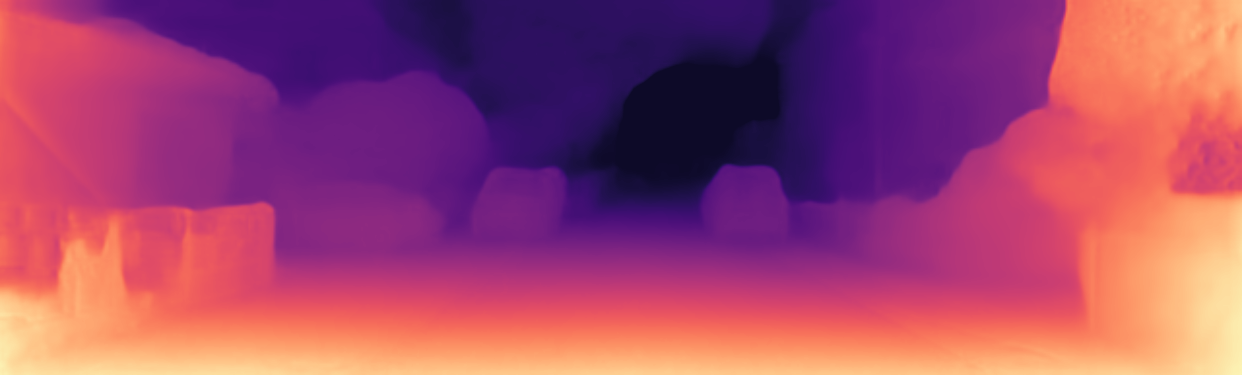}\hspace{\hspacing}%
    \includegraphics[width=\imgw]{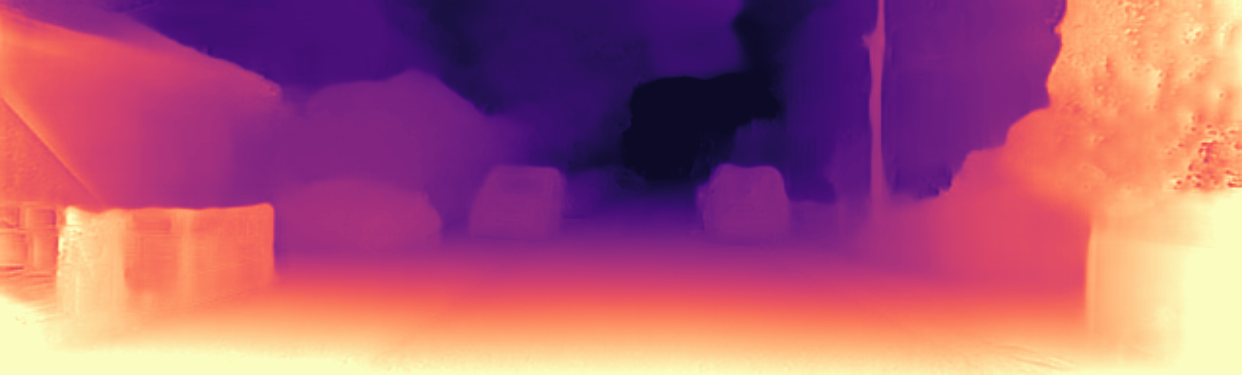}\hspace{\hspacing}%
    \includegraphics[width=\imgw]{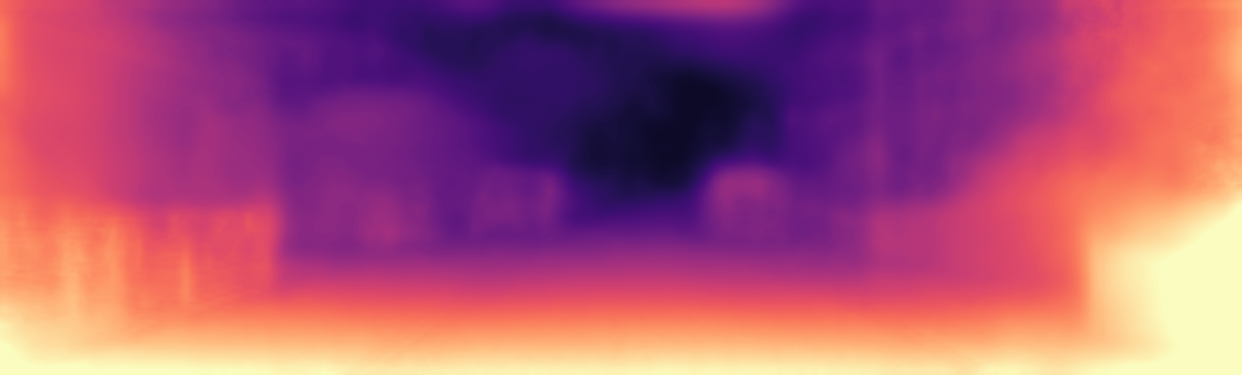}\hspace{\hspacing}%
    \includegraphics[width=\imgw]{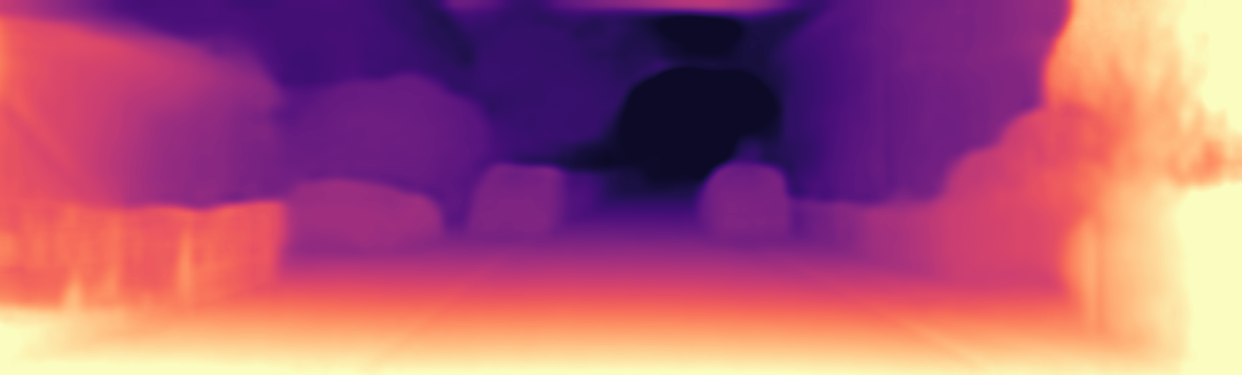}\hspace{\hspacing}%
    \vspace{\vspacing}


\includegraphics[width=\imgw]{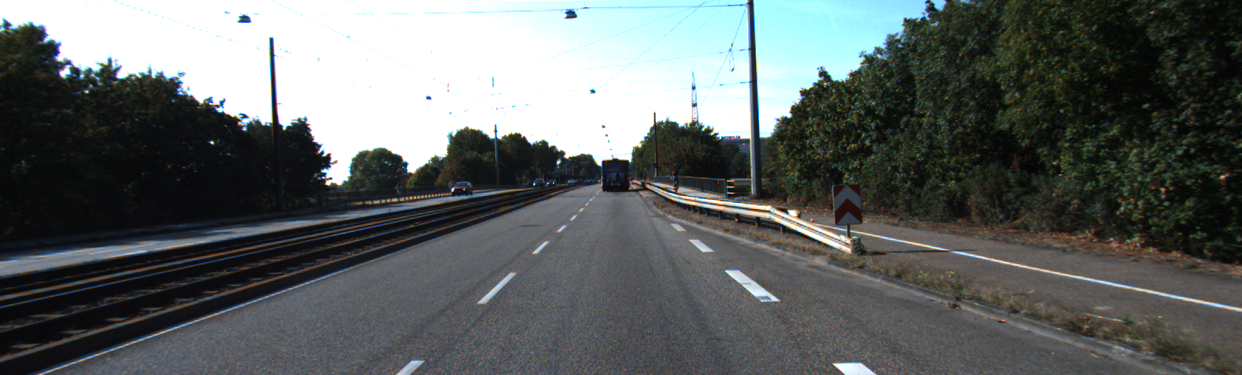}\hspace{\hspacing}%
    \includegraphics[width=\imgw]{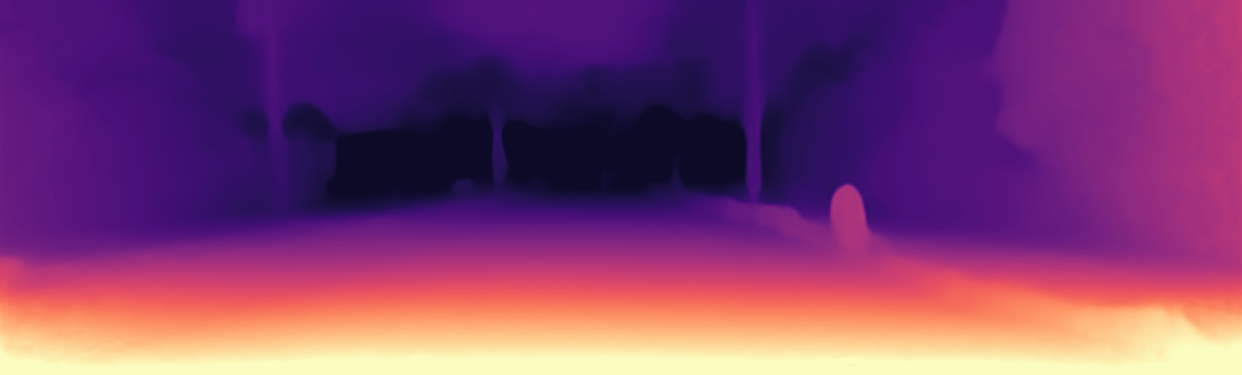}\hspace{\hspacing}%
    \includegraphics[width=\imgw]{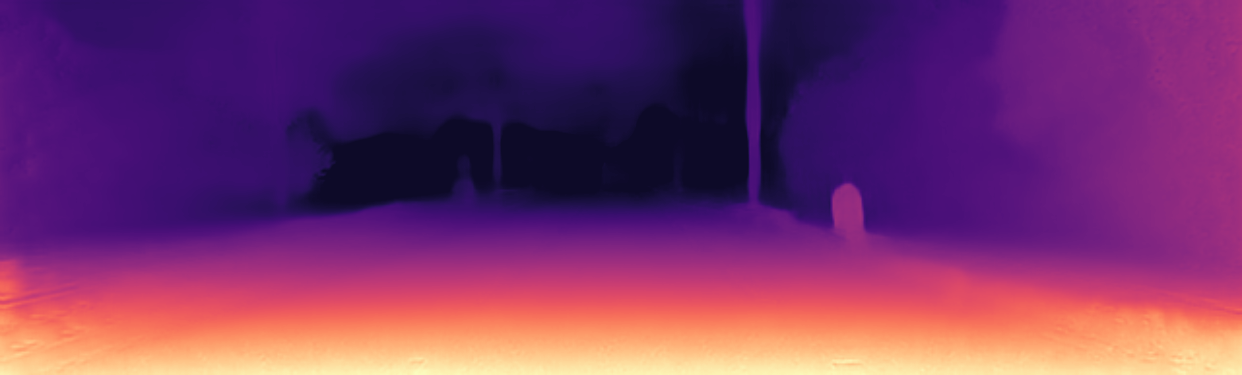}\hspace{\hspacing}%
    \includegraphics[width=\imgw]{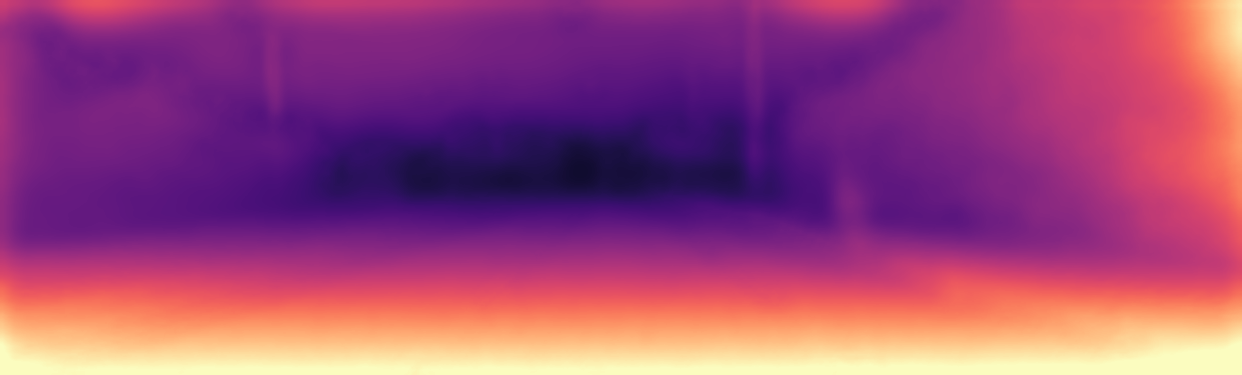}\hspace{\hspacing}%
    \includegraphics[width=\imgw]{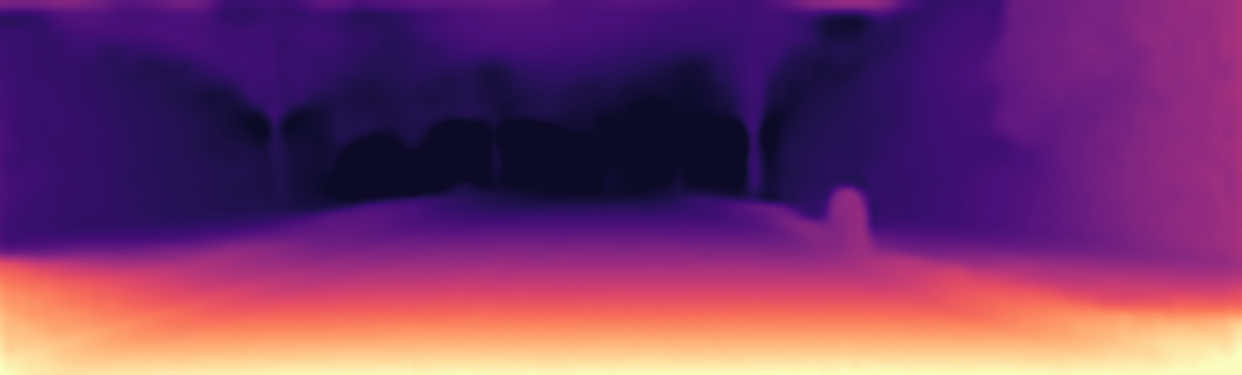}\hspace{\hspacing}%
    \vspace{\vspacing}

\includegraphics[width=\imgw]{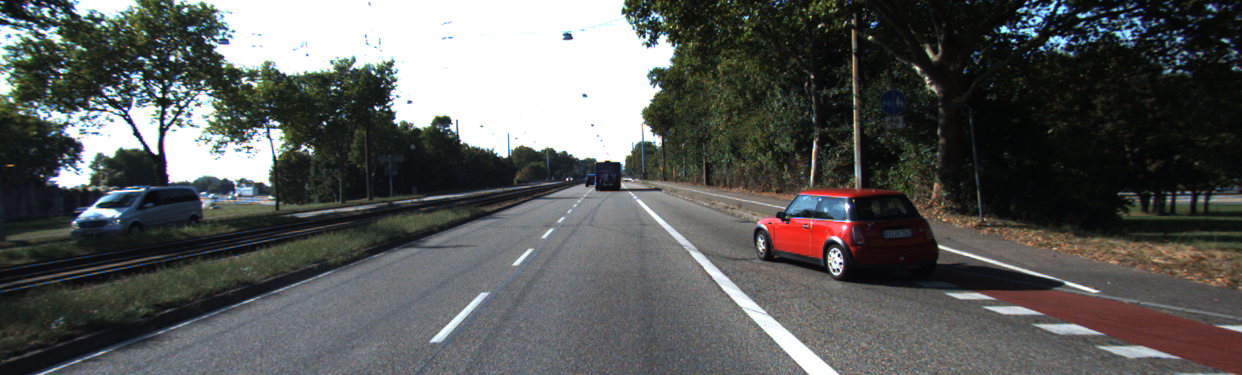}\hspace{\hspacing}%
    \includegraphics[width=\imgw]{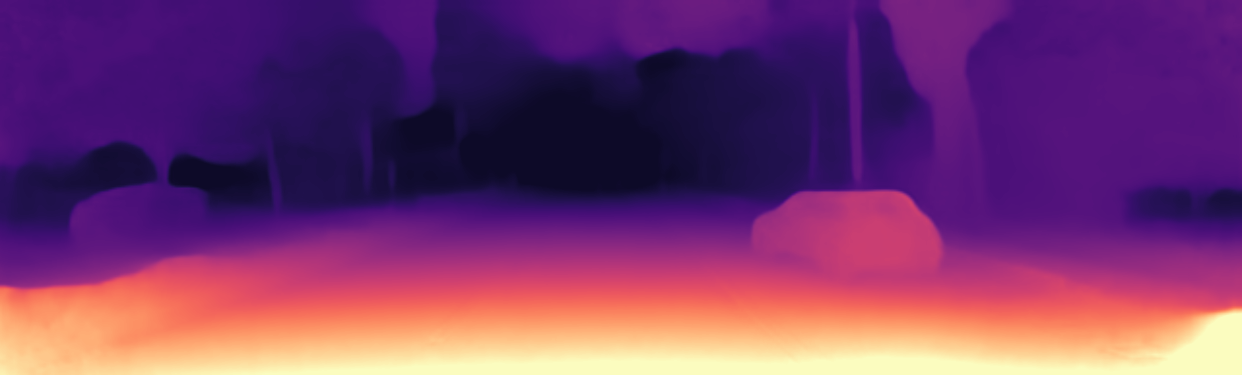}\hspace{\hspacing}%
    \includegraphics[width=\imgw]{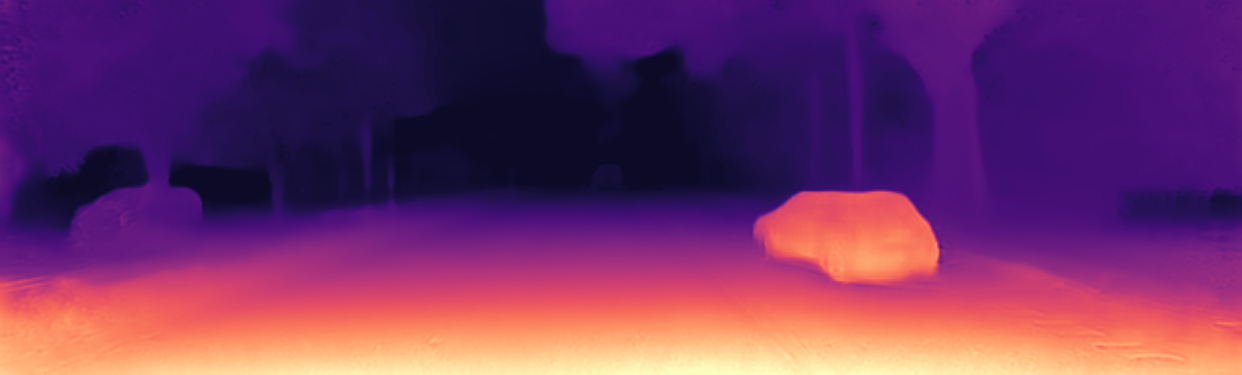}\hspace{\hspacing}%
    \includegraphics[width=\imgw]{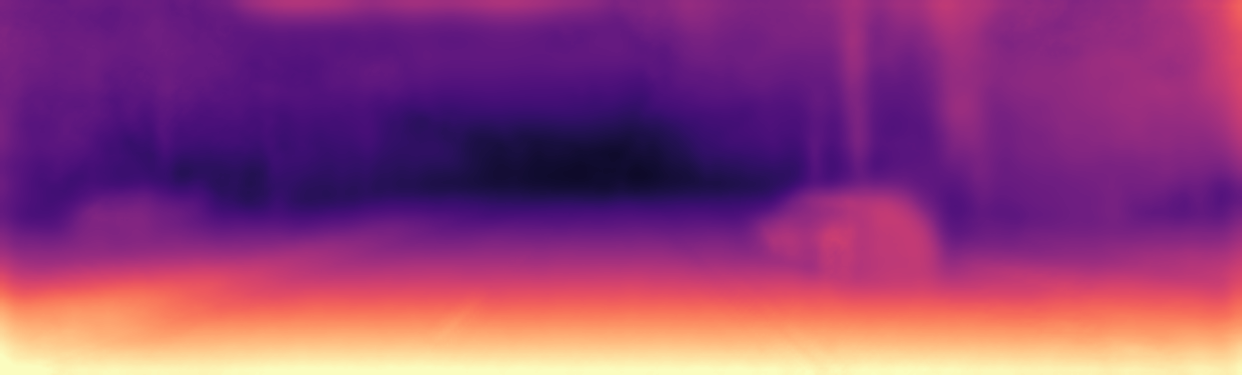}\hspace{\hspacing}%
    \includegraphics[width=\imgw]{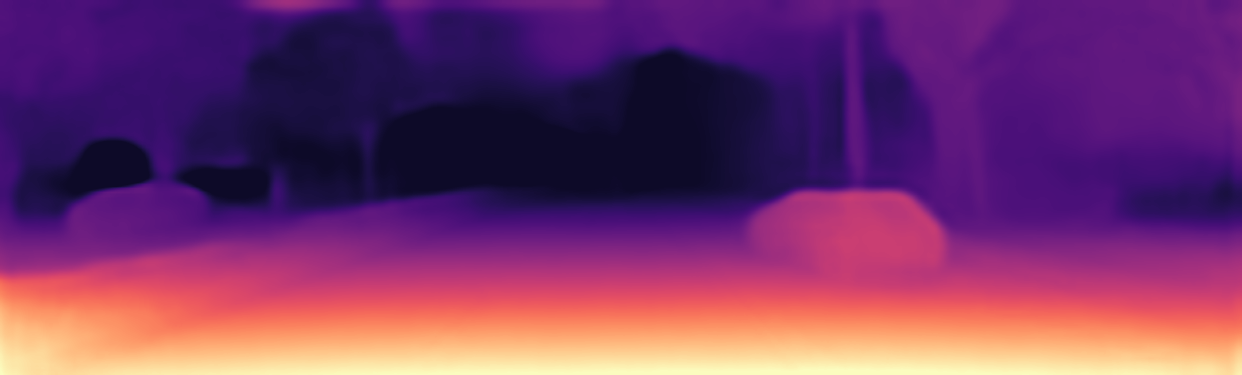}\hspace{\hspacing}%
    \vspace{\vspacing}




\includegraphics[width=\imgw]{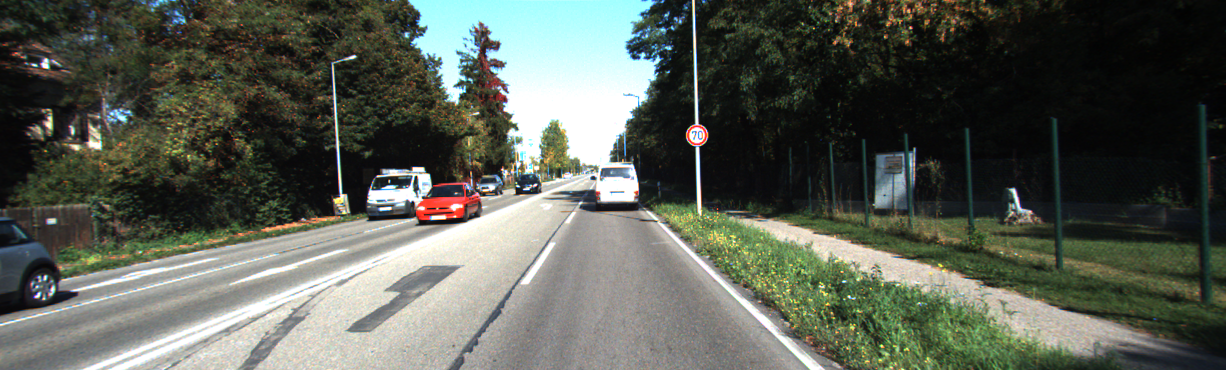}\hspace{\hspacing}%
    \includegraphics[width=\imgw]{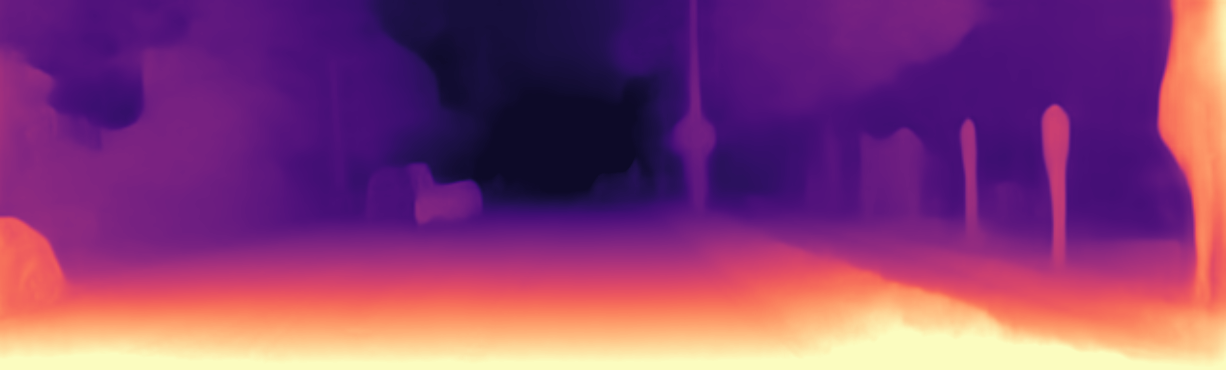}\hspace{\hspacing}%
    \includegraphics[width=\imgw]{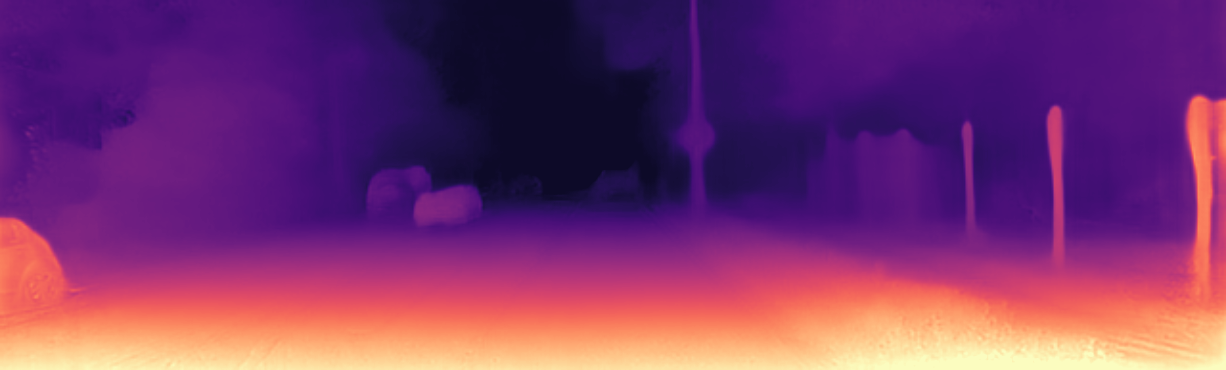}\hspace{\hspacing}%
    \includegraphics[width=\imgw]{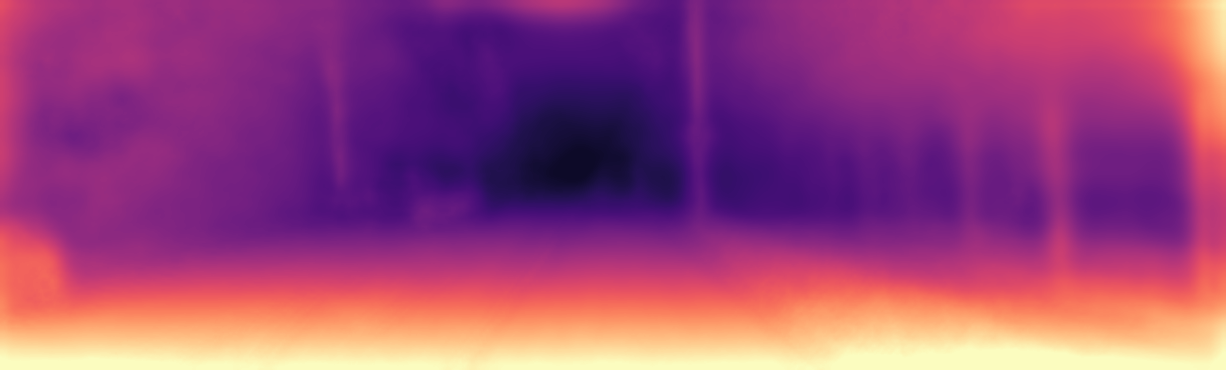}\hspace{\hspacing}%
    \includegraphics[width=\imgw]{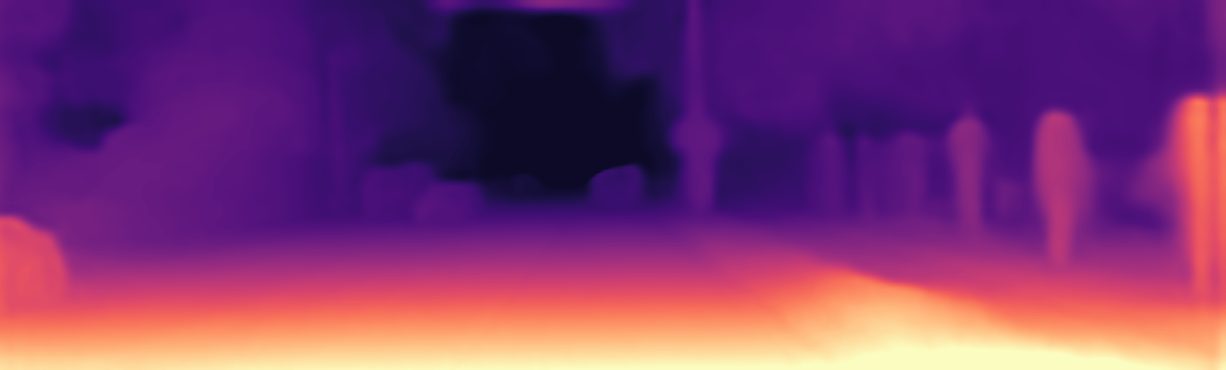}\hspace{\hspacing}%
    \vspace{\vspacing}

\includegraphics[width=\imgw]{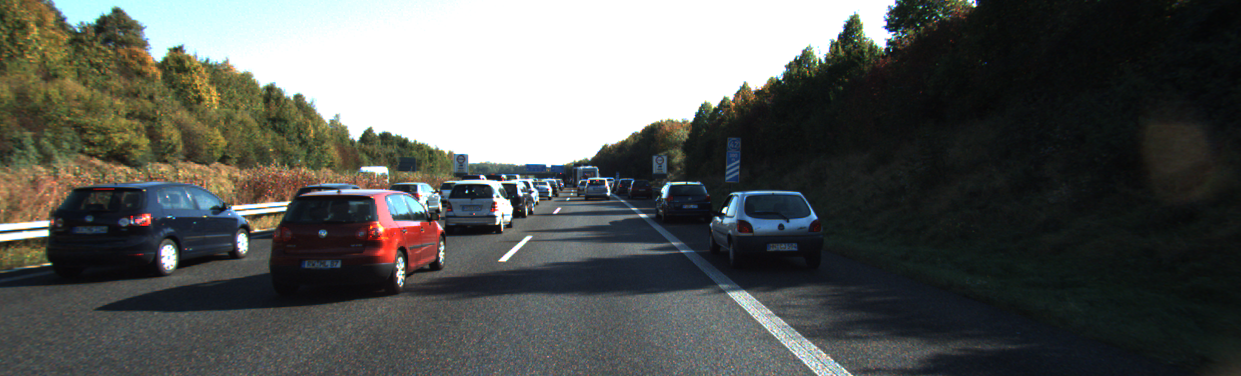}\hspace{\hspacing}%
    \includegraphics[width=\imgw]{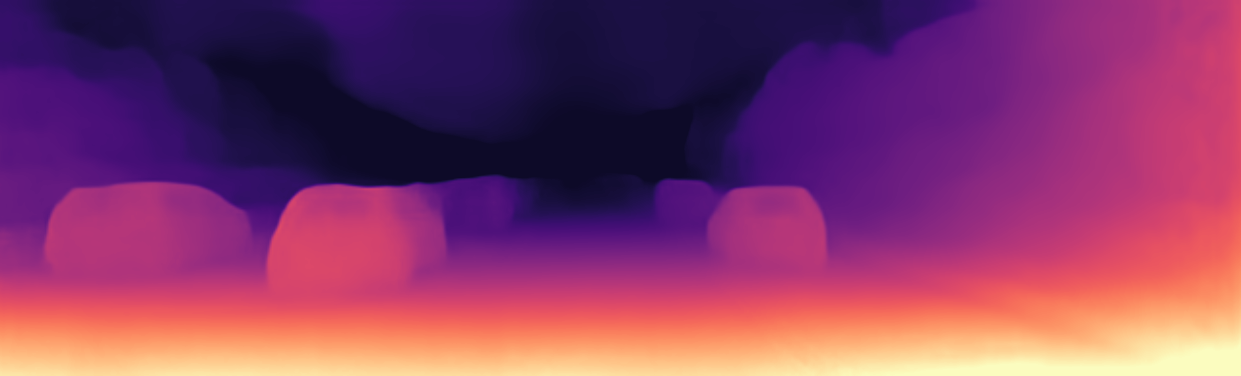}\hspace{\hspacing}%
    \includegraphics[width=\imgw]{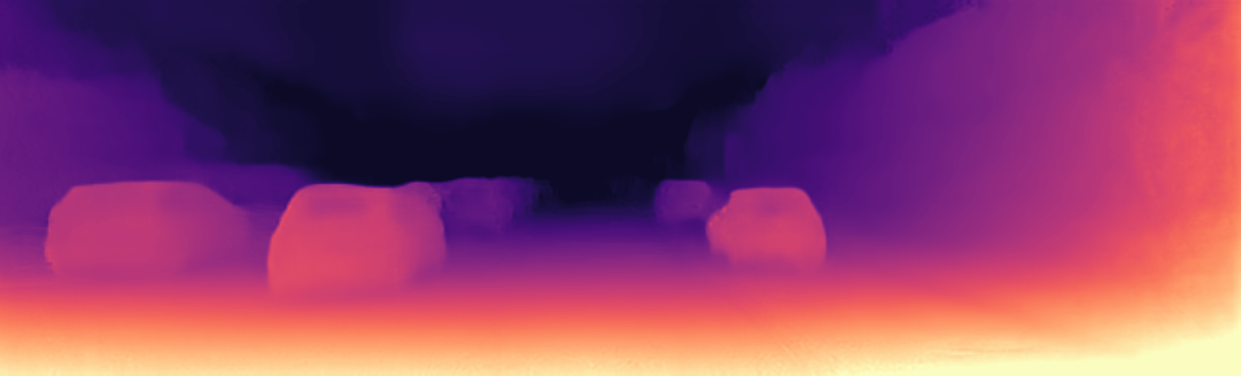}\hspace{\hspacing}%
    \includegraphics[width=\imgw]{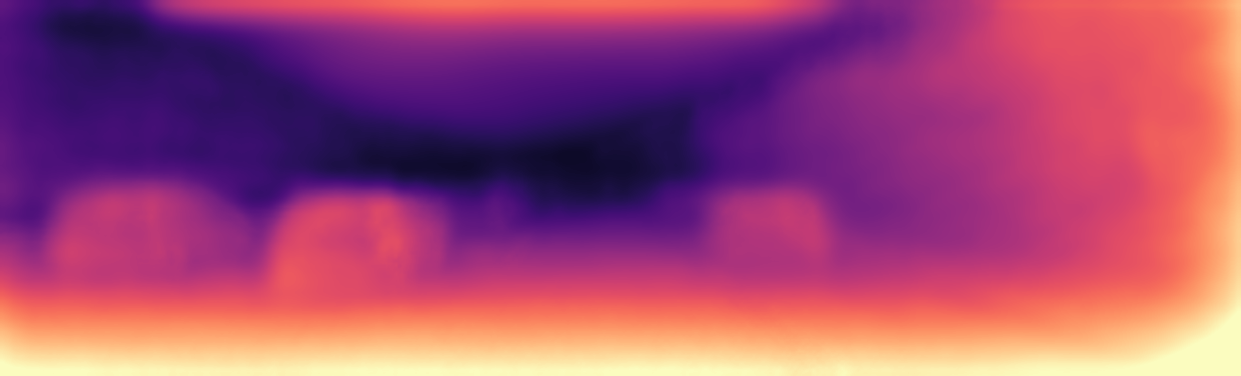}\hspace{\hspacing}%
    \includegraphics[width=\imgw]{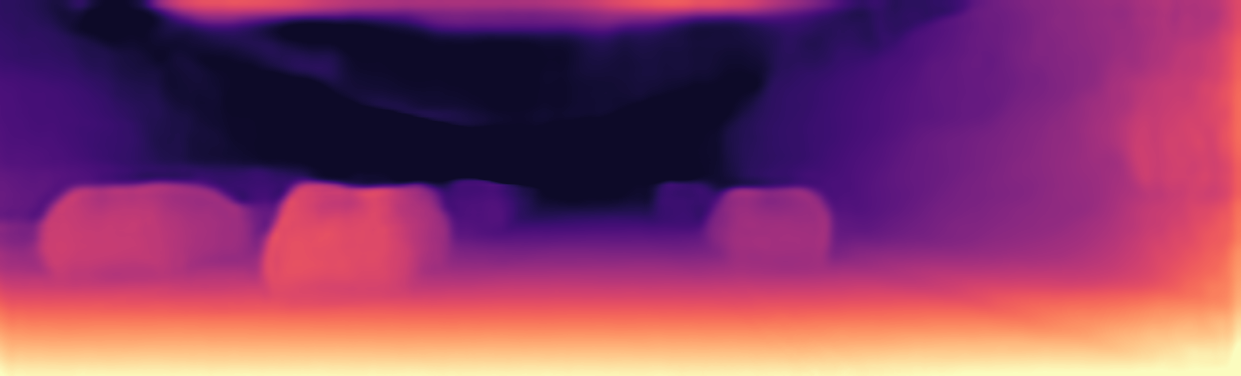}\hspace{\hspacing}%
    \vspace{\vspacing}


\includegraphics[width=\imgw]{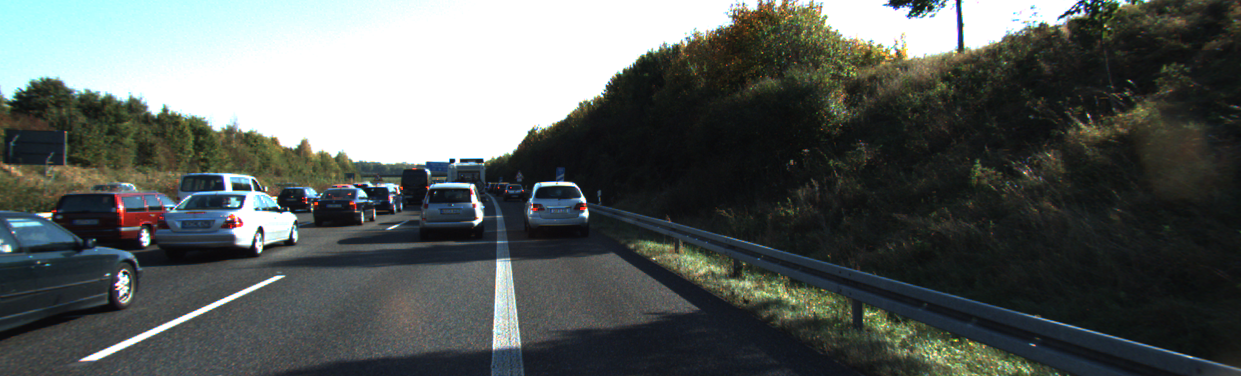}\hspace{\hspacing}%
    \includegraphics[width=\imgw]{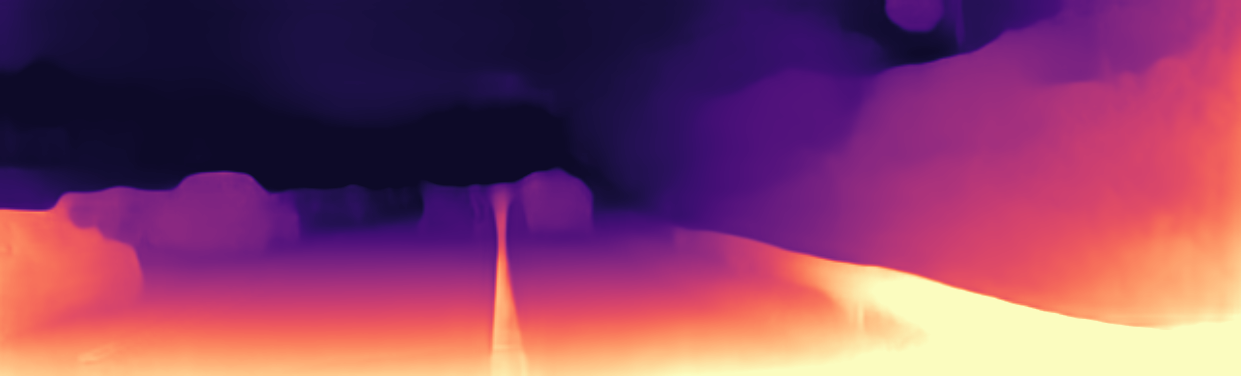}\hspace{\hspacing}%
    \includegraphics[width=\imgw]{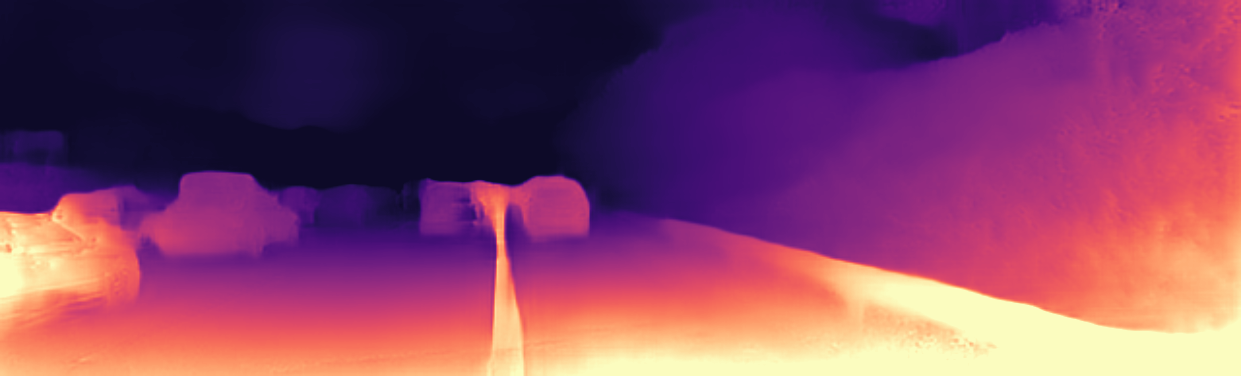}\hspace{\hspacing}%
    \includegraphics[width=\imgw]{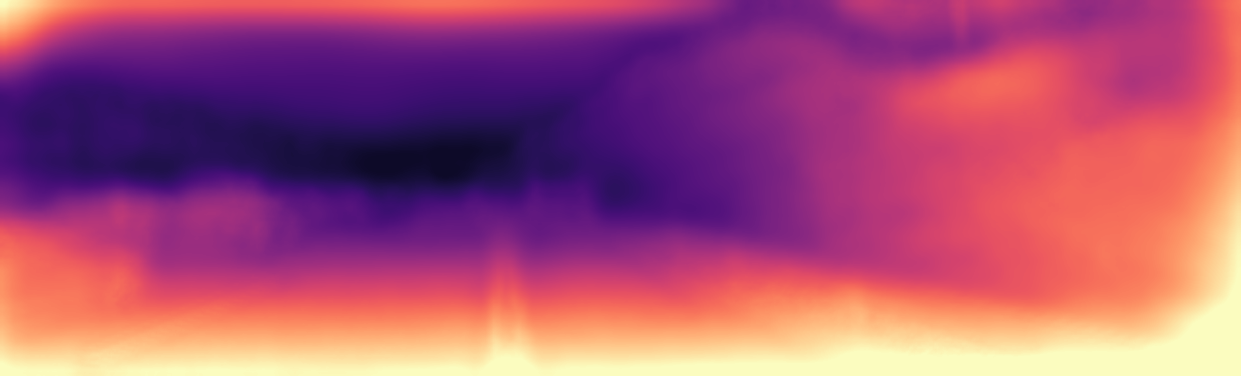}\hspace{\hspacing}%
    \includegraphics[width=\imgw]{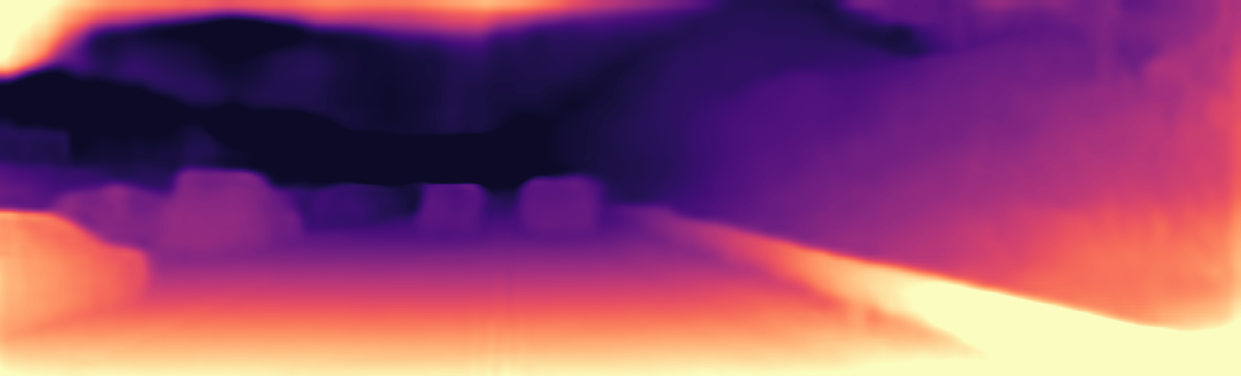}\hspace{\hspacing}%
    \vspace{\vspacing}

    
    \caption{{\bf Qualitative Results.} In each row, for an input image, we show the results of Monodepth2~\cite{Godard2019ICCV} and DNet~\cite{Xue2020IROS} and compare them to our models in the last two columns. \textit{Ours-Mono} refers to our ResNet18 model trained only on monocular sequences, and \textit{Ours-Stereo} refers to our ResNet50 model that was trained using additional stereo supervision.
    }
    \label{supp_fig:qualitative_results}
\end{figure*}